%% file: main.tex
\documentclass[10pt,twocolumn,letterpaper]{article}

\usepackage[pagenumbers]{cvpr}
\input{math_commands.tex}

\usepackage{float}
\usepackage{tabularx}
\usepackage{array}
\usepackage{algorithm}
\usepackage[noend]{algpseudocode}
\usepackage{wrapfig}
\usepackage{pifont}
\usepackage{placeins,needspace}
\definecolor{cvprblue}{rgb}{0.21,0.49,0.74}
\usepackage[breaklinks,colorlinks,allcolors=cvprblue]{hyperref}
\usepackage{booktabs}
\usepackage{multirow}
\usepackage{graphicx}

\title{Graph2Scene: Relation-Aware Text-to-Image Generation with Graph-Guided Diffusion Model}

\author{
Peizheng Li\quad Xin Ai\quad Guanghui Cheng\quad Zikang Dong\quad
Qiange Wang\thanks{Corresponding authors: Qiange Wang and Yanfeng Zhang.}\quad
Yanfeng Zhang\footnotemark[1]\\
School of Computer Science and Engineering, Northeastern University, China\\
{\tt\small \{lipz2, aix, chenggh, dongzikang\}@mails.neu.edu.cn}\quad
{\tt\small \{wangqg, zhangyf\}@mail.neu.edu.cn}
}

\hypersetup{pdftitle={Graph2Scene: Relation-Aware Text-to-Image Generation with Graph-Guided Diffusion Model},pdfauthor={Peizheng Li; Xin Ai; Guanghui Cheng; Zikang Dong; Qiange Wang; Yanfeng Zhang},pdfsubject={Research preprint}}

\newcommand{\ratioGood}[1]{\textsuperscript{\scriptsize$\times #1$}}
\newcommand{\ratioBad}[1]{\textsuperscript{\scriptsize$\times #1$}}
\newcommand{\ratioNeutral}[1]{\textsuperscript{\scriptsize$\times #1$}}
\newcommand{\ratioStrong}[1]{\textsuperscript{\textbf{\scriptsize$\times #1$}}}
\newcommand{\SectionEntry}[2]{\hangindent=1.5em\hangafter=1\hyperref[#1]{\ref*{#1}.\enspace #2}}

\begin{document}

\maketitle

\begin{abstract}
Text-to-image diffusion models still struggle with complex prompts containing many elements and interactions. Existing attention-control methods mainly focus on controlling where elements appear, but often ignore how objects interact with each other. As a result, relation tokens describing interactions between elements may assign high attention to irrelevant regions, leading to missing or incorrect interactions. In this paper, we propose Graph2Scene, a training-free framework that treats scene graphs as a core control signal for diffusion generation. It represents elements as nodes and their relations as edges, and translates these element-relation dependencies into selective attention guidance. It grounds each element to its corresponding region and guides relational interactions among the involved elements. To achieve appropriate relation control without over-constraining the generation process, Graph2Scene further adapts the control duration to scene complexity and progressively relaxes graph guidance during denoising. Experiments on T2I-CompBench and ConceptMix show clear improvements in compositional generation, especially for object relations. On ConceptMix, Graph2Scene improves relationship accuracy from 0.59 to 0.84 over the strongest FLUX.1-based regional-control baseline. Results on LongBench-T2I further show its advantage on long and complex prompts.
\end{abstract}

\section{Introduction}
Text-to-image diffusion models~\citep{rombach2022high,labs2025flux1,chen2024pixart} have made remarkable progress in generating realistic images from natural-language prompts. As model capabilities continue to advance, user demands are increasingly shifting toward complex prompts, where a target image may contain multiple objects with diverse attributes, spatial arrangements, and interactions~\citep{wu2024conceptmix, hoe2024interactdiffusion,zhou2025draw}. However, faithfully generating such complex scenes remains challenging, as the model must accurately capture the semantic dependencies among objects, attributes, and relations from long text prompts and translate them into the corresponding visual content.

In diffusion models, attention associates textual concepts with visual content during generation. Ideally, each text token should mainly influence the image region corresponding to its concept. However, in complex prompts, attention may spread to irrelevant regions, causing different concepts to interfere with each other.
We refer to this undesired behavior as attention leakage~\citep{wang2024compositional,dahary2024yourself,zhang2024easing}.
Attention leakage can lead to incorrect attributes, object interference, or missing elements during generation~\citep{chefer2023attend,wang2024compositional,rassin2023linguistic,marioriyad2024attention}.
Existing attention-control methods alleviate this issue by introducing spatial constraints, such as bounding boxes or region masks, to restrict object-related tokens to their corresponding regions~\citep{li2023gligen,wang2025ms,xie2023boxdiff,liu2022compositional,yang2024mastering}.

\begin{figure*}[t]
    \centering
    \includegraphics[width=0.82\textwidth]{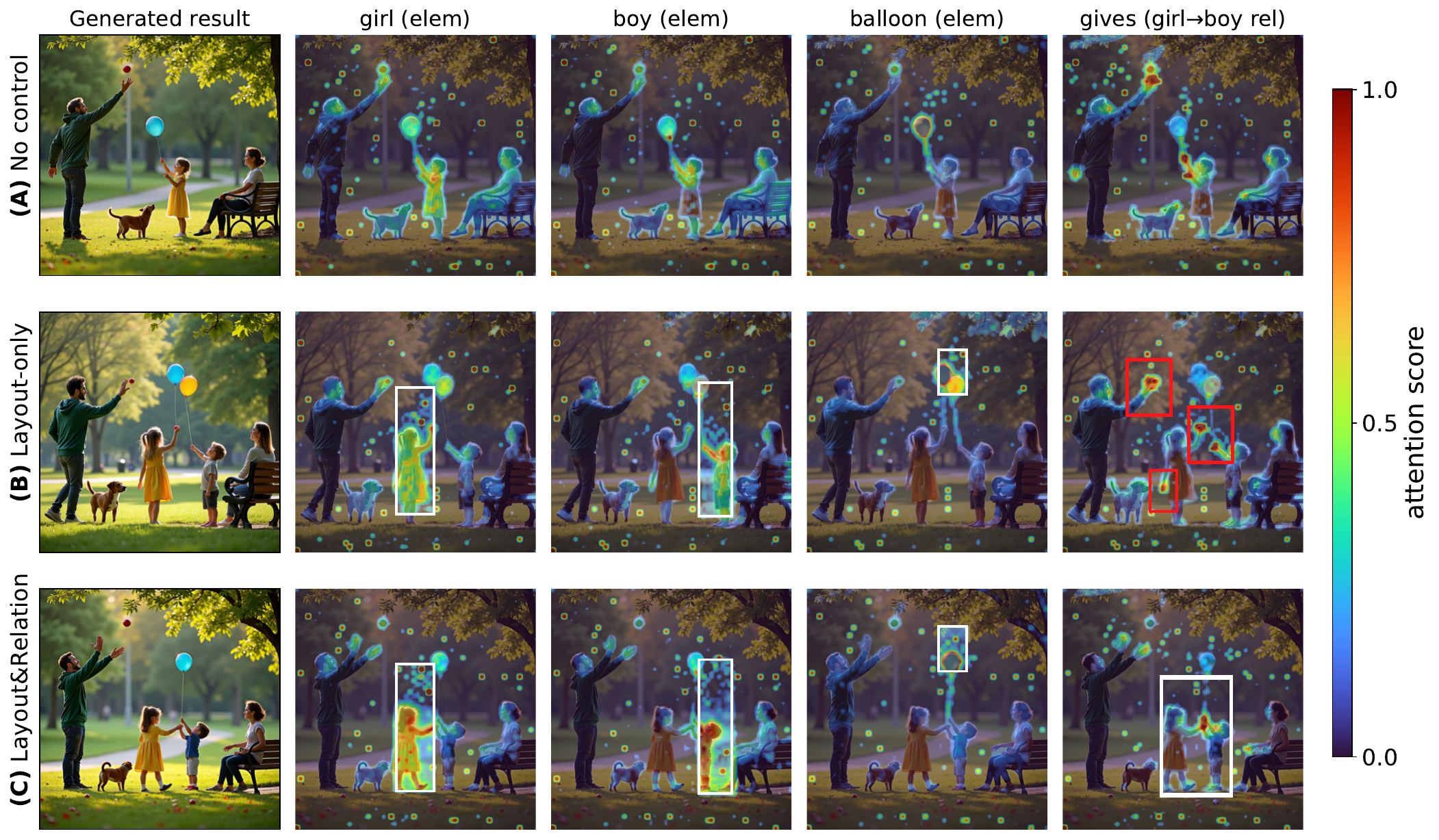}
    \caption{
    Cross-attention visualization on FLUX.1 for a selected interaction
    (\emph{a girl gives a balloon to a boy}) extracted from a complex prompt.
    Heatmaps show the aggregated attention of object and relation tokens.
    Rows compare (A) no control, (B) layout-only control, and (C) layout and relation control.
    White boxes indicate the regions to which the corresponding text tokens
    are constrained by the proposed control, while red boxes highlight
    the {attention leakage} phenomenon.}
    \label{fig:motivation}
\end{figure*}

Although existing methods can effectively place individual elements in their correct regions, they largely overlook attention leakage at the relation level. We conduct an experiment to verify that object-level layout control overlooks relation-level attention.
Figure~\ref{fig:motivation} compares (A) generation without attention control, (B) object-level layout control following~\citep{chen2024training} (C) the same layout control, while restricting relation tokens to focus only on the related elements and their interaction regions.
Without attention control (Figure~\ref{fig:motivation}(A)), the model fails to distinguish the regions of the girl and the boy, and the boy is missing. With layout control (Figure~\ref{fig:motivation}(B)), the girl, boy, and balloon are generated in the correct regions with proper attributes. However, the relation token \emph{gives} is not explicitly controlled, and its attention mainly focuses on the hand of an unrelated man rather than the girl, boy, and balloon involved in the interaction (as highlighted by the red box in Figure~\ref{fig:motivation}(B)). As a result, the intended giving relation is not correctly generated. By restricting \emph{gives} to attend to the related elements and their interaction region (Figure~\ref{fig:motivation}(C)), the intended relation becomes much clearer. This experiment demonstrates that object-level layout control alone is insufficient, and relation tokens also require explicit attention control.

Effectively controlling relations in complex scenes requires identifying which elements are involved in each interaction.
Scene graphs naturally provide such a structure by representing elements as nodes and interactions as edges, making the dependencies in complex prompts explicit. However, using scene graphs to guide complex image generation raises two key challenges. First, a scene graph only describes these dependencies and cannot directly guide a diffusion model. We need to determine how to translate its nodes and edges into effective attention operations. Second, the attention control mechanism should adapt to different denoising stages~\citep{chen2024delta,ma2024deepcache,lee2025beta}, while avoiding excessive intervention that may disturb the pretrained generation process.

To address these challenges, we propose \textbf{Graph2Scene}, a training-free framework that addresses relation control in complex text-to-image generation through two key designs. First, Graph2Scene introduces a graph-guided attention execution mechanism that converts scene graphs into structured attention operations. It explicitly models the correspondence between semantic concepts and visual regions: cross-attention associates relation concepts with their involved elements, while self-attention enables information propagation among graph-connected regions to reduce attention leakage and preserve object interactions. Second, Graph2Scene introduces a complexity-aware stage-wise execution strategy that estimates generation difficulty from attention patterns and adaptively determines when and how long graph control should be applied. This allows the framework to provide necessary guidance for challenging scenes while preserving the native generation ability of the pretrained model.

In summary, our contributions are as follows:
\begin{itemize}[leftmargin=*,itemsep=2pt,topsep=4pt]

\item We identify that faithful generation of complex scenes requires explicit coordination of attention among interacting objects, beyond the object grounding and spatial control emphasized by existing methods.

\item
Based on this observation, we propose \textbf{Graph2Scene}, a training-free framework that translates scene graphs into relation-aware attention control. It compiles element-relation dependencies into selective cross- and self-attention guidance and adaptively executes such guidance along the denoising trajectory according to scene complexity.

\item Extensive experiments on T2I-CompBench, ConceptMix, and LongBench-T2I show that Graph2Scene consistently improves compositional generation, especially on complex relation-rich scenes, improving relationship accuracy on ConceptMix from 0.42 to 0.84 compared with the FLUX.1 model.

\end{itemize}

\section{RELATED WORK}
\subsection{Compositional Diffusion Generation.}
Text-to-image diffusion models\citep{rombach2022high,labs2025flux1,chen2024pixart} can generate high-quality images from text. However, they often fail on compositional prompts with multiple objects, attributes, and relations. Common errors include missing objects, wrong attribute binding, and incorrect relations\citep{dahary2024yourself,marioriyad2024attention}. This has motivated many studies on compositional diffusion generation.

One line of work injects external layout signals into the attention. GLIGEN~\citep{li2023gligen} adds gated attention layers that take grounding boxes as input, and MS-Diffusion~\citep{wang2025ms} trains an adapter
that binds reference subjects to given regions. These methods pre-train small modules, and then steer where each object appears.

A second line of work directly constrains or optimizes the existing attention maps at inference time, without adding modules.Attend-and-Excite~\citep{chefer2023attend} strengthens the cross-attention responses of subject tokens during inference, preventing important concepts from being neglected in the generated image. Composable Diffusion~\citep{liu2022compositional} separately predicts the denoising direction for each concept and combines them at every diffusion step to generate all concepts together. Regional Prompt~\citep{chen2024training} splits the attention by region so each prompt attends to its own area. LLM-Blueprint~\citep{gani2024llm}
and RPG~\citep{yang2024mastering} further use an LLM to decompose complex prompts and plan object layouts before applying region-level generation or attention control. These methods reshape attention per object, but they treat each object on its own. Although these methods\citep{hoe2024interactdiffusion,mei2026semantic} consider interactions, they mainly focus on generating specific interaction patterns in simple scenes.
They rely on predefined HOI categories or learned interaction representations, and cannot generalize to diverse relations among multiple elements in complex scenes.

Overall, existing methods mainly improve compositional generation through object grounding, attribute binding, or layout control. Their guidance is largely defined at the object level and does not explicitly model the interactions among multiple elements. As scene complexity increases, this object-wise control is insufficient to capture relation-dependent generation, leading to incorrect or missing interactions even when individual objects are properly placed.

\subsection{Scene Graphs for Diffusion Models}
Scene graphs are also used to represent objects and relations in complex scenes. Existing methods mainly differ in how the graph is converted into a generation condition. SG2Im~\citep{johnson2018image} uses a GCN to encode the graph and
predict object layouts. SGDiff~\citep{yang2022diffusion} learns graph or relation embeddings. CanonicalSg2Im~\citep{herzig2020learning} learns canonical graph representations to handle
semantically equivalent graphs. SG-Adapter~\citep{shen2025scene} uses graph triplets to refine text embeddings.
R3CD~\citep{liu2024r3cd} further models relations with a graph transformer and relation-aware attention objectives.

Most existing scene graph-based methods focus on scene graph-to-image generation, where scene graphs are directly used as inputs to graph-conditioned generators or trained graph encoders. In contrast, Graph2Scene does not train a new graph-conditioned generation model, but explores how to execute scene graph constraints on a pretrained text-to-image diffusion model through attention control. Therefore, in the main experiments, we mainly compare with existing compositional generation methods built upon pretrained T2I models, while providing additional comparisons with scene graph-based generation methods in Section~C of the supplementary material.

\section{Graph2Scene}
\begin{figure*}[t]
    \centering
    \includegraphics[width=0.83\linewidth]{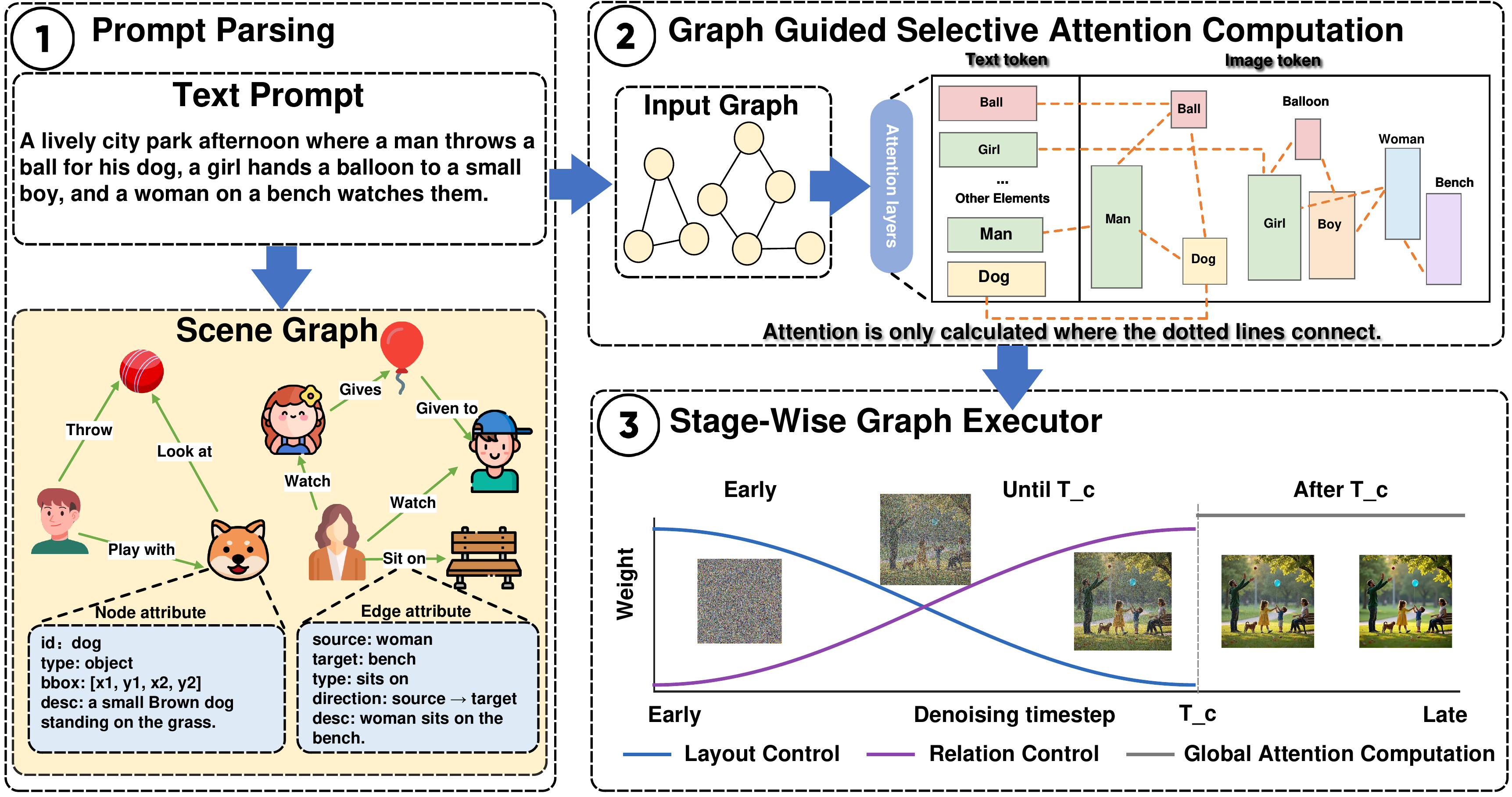}
    \caption{Workflow of Graph2Scene.}
    \label{fig:overview}
\end{figure*}

As shown in Figure.~\ref{fig:overview}, Graph2Scene consists of three main components. First, we construct a scene graph from a complex text prompt. Each element node contains its semantic description and predicted layout, while each relation edge connects the corresponding elements involved in an interaction. Second, we compile the scene graph into attention masks for the frozen Diffusion Transformer. In cross-attention, the masks control the association between text tokens and image regions, so that element descriptions attend to their target regions and relation tokens are bound to the regions of the elements involved in each interaction. In self-attention, the masks restrict information exchange according to the graph structure, allowing communication within each element region and between graph-connected regions, which further reduces attention leakage across unrelated regions. Third, we apply these graph-based controls along the denoising trajectory with a stage-wise schedule. The control duration is determined by the estimated scene complexity, allowing the framework to adjust the execution process across different denoising stages.

\subsection{Hierarchical Scene Graph Construction}
\label{sec:graph_construction}
Graph2Scene represents a complex prompt as a scene graph of elements and their relations, making the structural dependencies in the prompt explicit. Specifically, we use a vision--language model (VLM) to parse the prompt into a hierarchical representation with a global scene and individual element nodes. Each element node stores its semantic description and predicted layout information, including its bounding box. Each edge connects two element nodes and specifies their relation, such as holding or next to. The graph explicitly represents where each element should appear and how different elements should interact, providing structured guidance for attention computation during the subsequent denoising process. Construction examples and prompt templates are provided in Sections~A and~D of the supplementary material.

\subsection{Graph guided selective attention computation}
\label{sec:compile}
To ensure that relational concepts interact with their corresponding elements, Graph2Scene introduces a customized attention mask mechanism for the MM-DiT backbone. The mask is constructed from the scene graph structure, including the topology of nodes and edges, as well as the layout and semantic information associated with each node and edge.

\begin{figure*}[t]
    \centering
    \includegraphics[width=0.84\linewidth]{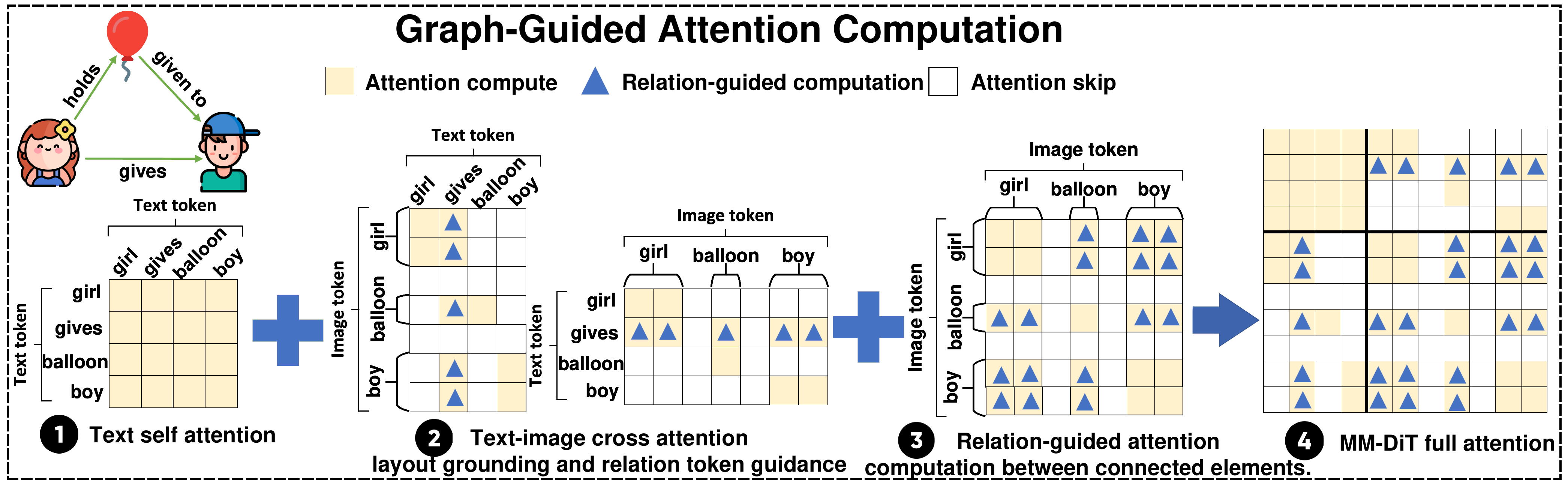}
    \caption{Illustration of graph-guided attention computation. Cross-attention associates element and relation tokens with the corresponding image regions, while self-attention allows information exchange within each region and between graph-connected regions.}
    \label{fig:attention}
\end{figure*}

As illustrated in Figure~\ref{fig:attention}, Graph2Scene converts the scene graph into structured attention computation through a customized attention mask, which regulates information flow in the MM-DiT backbone. Graph2Scene performs graph-guided attention through two operations: text-image cross-attention and relation-guided self-attention. The \textbf{text-image cross-attention computation} (\ding{203}) associates text tokens with image tokens. Element tokens are connected to their corresponding image regions, while relation tokens are connected to the regions of all participating elements. This provides layout grounding and binds each relation to its related elements. The \textbf{relation-guided self-attention computation} (\ding{204}) controls information exchange among image tokens. Tokens within the same element region can interact with each other, while tokens from different regions can exchange information only when their corresponding elements are connected in the scene graph. Together with text self-attention (\ding{202}), these operations define the graph-guided attention computation, which is integrated into the full MM-DiT attention (\ding{205}).

Graph2Scene translates the scene graph into a customized attention mask $M_{\mathcal{G}}$, which regulates information flow during attention computation. The mask consists of two parts: a cross-attention mask $M_{\mathrm{cross}}$, which controls the interaction between text tokens and image regions, and a self-attention mask $M_{\mathrm{self}}$, which regulates information exchange among image tokens. Given a scene graph $\mathcal{G}=(\mathcal{V},\mathcal{E})$, each node $v\in\mathcal{V}$ represents an element with its associated text token set $\mathcal{T}_v$ and spatial region $\mathcal{R}_v$ predicted by the VLM. Each edge $e\in\mathcal{E}$ represents a relation with token set $\mathcal{T}_e$, where $\mathcal{V}_e$ denotes the set of element nodes involved in this relation. We denote the image tokens corresponding to the region of node $v$ as $\mathcal{X}_v$. Based on these graph associations, the attention mask determines the allowed token interactions during attention computation.

\paragraph{Cross-attention mask.}
The cross-attention mask controls the interaction between text tokens and image regions in both directions (\ding{203}).
For each element node, its associated text tokens are connected to the corresponding image region.
For each relation edge, the relation tokens are connected to the image regions of all participating nodes, allowing relational concepts to interact with the elements involved in the relation.
Accordingly, we apply the same graph-guided connectivity to both text-to-image and image-to-text attention:
\begin{equation}
\begin{aligned}
& M_{\mathrm{cross}}^{\mathrm{T}\rightarrow\mathrm{I}}(i,j)
= M_{\mathrm{cross}}^{\mathrm{I}\rightarrow\mathrm{T}}(j,i)
\\
& \quad =
\begin{cases}
0,
& \begin{aligned}[t]
  & \text{if } (t_i \in \mathcal{T}_v
    \land x_j \in \mathcal{X}_v) \\
  & \quad \lor (t_i \in \mathcal{T}_e
    \land x_j \in
    \bigcup\nolimits_{v \in \mathcal{V}_e}\mathcal{X}_v),
  \end{aligned}
\\
-\infty, & \text{otherwise}.
\end{cases}
\end{aligned}
\end{equation}
\paragraph{Self-attention mask.}
The self-attention mask controls information exchange among image tokens (\ding{204}). In self-attention, each element region is first allowed to interact with itself to preserve its local visual consistency. Beyond self-region attention, we only enable interactions between regions whose corresponding elements are connected by a relation edge in the scene graph. This design maintains independent object representations while facilitating visual information exchange among relationally connected elements. Formally, the self-attention mask is defined as:
\begin{equation}
M_{\mathrm{self}}(i,j) =
\begin{cases}
0,
& \begin{aligned}[t]
  & \text{if } (\exists v\in\mathcal{V}: \\
  & \quad x_i,x_j\in\mathcal{X}_v) \\
  & \lor (\exists e\in\mathcal{E},\
    u,v\in\mathcal{V}_e: \\
  & \quad x_i\in\mathcal{X}_u,\
    x_j\in\mathcal{X}_v),
  \end{aligned} \\
-\infty, & \text{otherwise}.
\end{cases}
\end{equation}
The overall attention computation with the graph-guided mask is formulated as:
\begin{equation}
\mathrm{Attention}(Q,K,V;M_{\mathcal{G}})
=
\mathrm{Softmax}\left(
    \frac{QK^{T}}{\sqrt{d}} + M_{\mathcal{G}}
\right)V .
\end{equation}

\subsection{Stage-wise Graph Execution}
\label{sec:exec}

Graph-guided attention control should be adaptive rather than applied with a fixed strategy for all prompts.
First, continuous attention control throughout generation may harm the model's native generation ability, so the control should be stopped in time once sufficient guidance has been provided.
Second, relation control should follow layout control, since diffusion models typically establish object layouts before refining semantic relations and visual details~\citep{lee2025beta,pan2025t}.

\begin{figure}[!tb]
    \centering
    \includegraphics[width=0.95\linewidth]{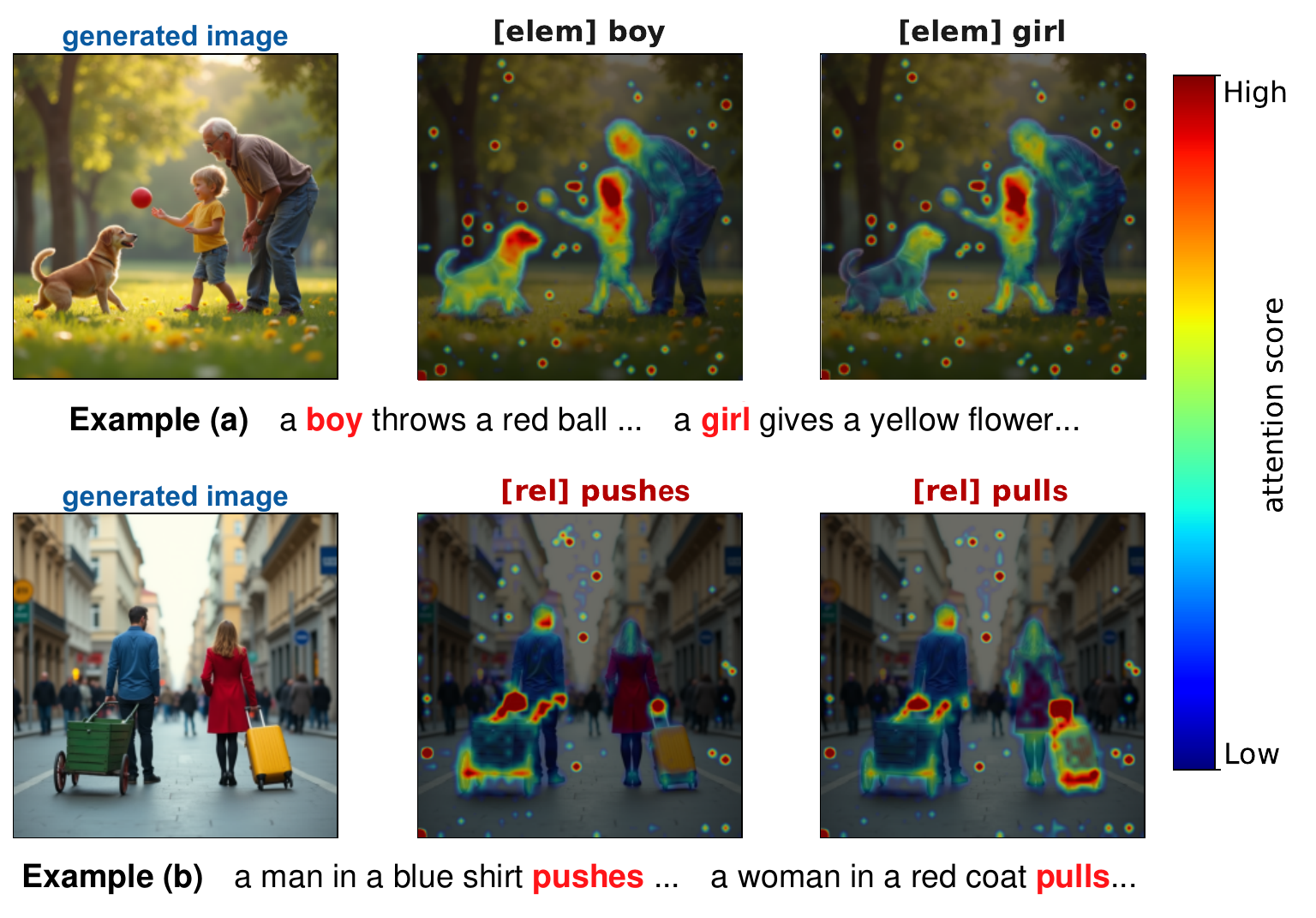}
    \caption{Visualization of two typical attention failures. Example (a) shows attention overlap between the \textit{boy} and \textit{girl} tokens, while example (b) shows attention leakage of the relation tokens \textit{pushes} and \textit{pulls}.}
    \label{fig:leakage}
\end{figure}

We monitor the attention distributions of specific text tokens during generation to guide the adaptive adjustment of control strength. These attention distributions reveal two typical generation problems: attention overlap and attention leakage.
Attention overlap occurs when the high-score attention of different tokens concentrates on largely the same spatial regions.
Attention leakage occurs when the high-score attention distribution of a token extends beyond the region of its corresponding concept. Figure~\ref{fig:leakage} illustrates both cases: in example (a), the attention maps of \emph{boy} and \emph{girl} overlap heavily, causing the girl to be missing, while in example (b), the relation tokens \emph{pushes} and \emph{pulls} fail to focus on their corresponding regions, leading to incorrect relation generation.

\begin{table*}[!t]
\centering
\caption{
Quantitative comparison on T2I-CompBench using VLM-based
accuracies and the official benchmark metrics.
Higher scores indicate better performance.
}
\label{tab:t2i_compbench}
\setlength{\tabcolsep}{3.2pt}
\begin{tabular*}{\linewidth}{@{\extracolsep{\fill}}lccccccccc@{}}
\toprule
\multirow{3}{*}{Method}
& \multicolumn{3}{c}{VLM-based Evaluation $\uparrow$}
& \multicolumn{6}{c}{Official Evaluation $\uparrow$} \\
\cmidrule(lr){2-4}
\cmidrule(lr){5-10}
& \multirow{2}{*}{Acc$_{\text{exist}}$}
& \multirow{2}{*}{Acc$_{\text{attribute}}$}
& \multirow{2}{*}{Acc$_{\text{relationship}}$}
& \multicolumn{3}{c}{Attribute Binding}
& \multicolumn{2}{c}{Object Relationships}
& \multirow{2}{*}{Complex} \\
\cmidrule(lr){5-7}
\cmidrule(lr){8-9}
& & & & Color & Shape & Texture & Spatial & Non-spatial & \\
\midrule
FLUX.1 & 0.79 & 0.66 & 0.60 & 0.7763 & 0.6370 & 0.8057 & 0.3499 & 0.3057 & 0.3858 \\
GLIGEN & 0.67 & 0.43 & 0.62 & 0.1762 & 0.3428 & 0.4926 & 0.2715 & 0.3090 & 0.3020 \\
Composable & 0.61 & 0.41 & 0.44 & 0.2503 & 0.3776 & 0.5324 & 0.0000 & 0.2822 & 0.2867 \\
Attend-and-Excite & 0.64 & 0.42 & 0.28 & 0.3638 & 0.4320 & 0.6444 & 0.0789 & 0.3082 & 0.2785 \\
MS-Diffusion & 0.81 & 0.51 & 0.53 & 0.4704 & 0.6392 & 0.6530 & 0.2488 & 0.3094 & 0.3420 \\
LLM-Blueprint & 0.78 & 0.50 & 0.64 & 0.3753 & 0.4788 & 0.6714 & 0.3317 & 0.3048 & 0.2799 \\
RPG & 0.80 & 0.55 & 0.70 & 0.5303 & 0.6153 & 0.6598 & 0.3481 & 0.3088 & 0.3248 \\
Regional Prompt & 0.88 & 0.69 & 0.73 & 0.8169 & 0.6742 & 0.7585 & 0.5483 & 0.3105 & 0.4196 \\
\midrule
\textbf{Graph2Scene (Ours)} & \textbf{0.94} & \textbf{0.80} & \textbf{0.89} & \textbf{0.8551} & \textbf{0.6982} & \textbf{0.8364} & \textbf{0.5816} & \textbf{0.3149} & \textbf{0.4374} \\
\bottomrule
\end{tabular*}
\end{table*}

\paragraph{Control Strength Adjustment.}
Based on these attention behaviors, we first perform a warm-up generation without additional guidance and collect the attention distributions of the selected text tokens. We then measure the severity of attention overlap and attention leakage from these attention maps. For prompts with more severe attention overlap or leakage, Graph2Scene applies stronger and longer graph guidance.
The detailed calculation is provided in Section~E of the supplementary material.

\paragraph{Control Selection Strategy.}
Graph2Scene progressively shifts the control focus from layout guidance to relation guidance during denoising. This design follows how diffusion models gradually form the layout first and refine semantics and interactions later. Since this transition is gradual and continuous~\citep{zhang2025blockdance}, we linearly decrease the layout guidance weight from 1 to 0 and increase the relation guidance weight from 0 to 1 over the determined \mbox{control duration.}

\section{EXPERIMENT}

\subsection{EXPERIMENT SETTING}

\begin{table*}[!t]
\centering
\caption{Quantitative comparison on ConceptMix.}
\label{tab:conceptmix}
\begin{tabular*}{\linewidth}{@{\extracolsep{\fill}}lcccccc@{}}
\toprule
Method & Acc$_{\text{exist}}$ $\uparrow$ & Acc$_{\text{attribute}}$ $\uparrow$ & Acc$_{\text{relationship}}$ $\uparrow$ & CLIPScore $\uparrow$ & HPSv2 $\uparrow$ & PickScore $\uparrow$ \\
\midrule
FLUX.1 & 0.75 & 0.31 & 0.42 & 24.82 & 0.27 & 22.31 \\
GLIGEN & 0.28 & 0.04 & 0.06 & 19.28 & 0.21 & 20.04 \\
Composable & 0.17 & 0.17 & 0.04 & 16.51 & 0.19 & 19.44 \\
Attend-and-Excite & 0.33 & 0.06 & 0.10 & 20.20 & 0.22 & 20.50 \\
MS-Diffusion & 0.47 & 0.06 & 0.11 & 21.76 & 0.28 & 21.97 \\
LLM-Blueprint & 0.29 & 0.06 & 0.11 & 19.66 & 0.21 & 20.09 \\
RPG & 0.40 & 0.07 & 0.09 & 21.88 & 0.27 & 21.93 \\
Regional Prompt & 0.80 & 0.64 & 0.59 & 25.91 & 0.29 & 21.76 \\
\midrule
Graph2Scene & \textbf{0.88} & \textbf{0.74} & \textbf{0.84} & \textbf{26.45} & \textbf{0.29} & \textbf{22.45} \\
\bottomrule
\end{tabular*}
\end{table*}
\textbf{Datasets.}
We evaluate Graph2Scene on two complementary benchmarks covering different levels of scene complexity. \textbf{T2I-CompBench}~\citep{huang2023t2i} evaluates compositional generation, including attribute binding and object relationships, using relatively concise prompts. For complex scenes, motivated by prior efforts to construct semantically challenging prompts~\citep{niu2026wise,zhang2025unified}, we adapt the concept-mixing strategy of ConceptMix~\citep{wu2024conceptmix} to construct prompts with controllable compositional complexity. Specifically, we extend the concept range to $K\in[8,20]$ and introduce an additional \emph{interaction} dimension to increase the density of inter-object relationships.
We additionally report qualitative results on the long-prompt benchmark \textbf{LongBench-T2I}~\citep{zhou2025draw} in Section~G of the supplementary material.

\textbf{Baselines.} We compare Graph2Scene with representative compositional and controllable generation methods, including \textbf{Composable Diffusion}~\citep{liu2022compositional},
\textbf{GLIGEN}~\citep{li2023gligen}, \textbf{Attend-and-Excite}~\citep{chefer2023attend}, \textbf{MS-Diffusion}~\citep{wang2025ms},
\textbf{RPG}~\citep{yang2024mastering}. Since these methods are built on different diffusion backbones, we follow their
official implementations and recommended settings. We additionally include \textbf{FLUX.1} and the FLUX.1-based \textbf{Regional Prompt} for direct same-backbone comparison~\citep{chen2024training}. Runtime comparisons with Regional Prompt across different prompt lengths and image resolutions are reported in Section~F of the supplementary material.

\textbf{Metrics.}

We adopt complementary evaluation protocols for scenes of varying complexity. For simple compositional generation, we follow the official evaluation protocol of T2I-CompBench to ensure comparability with existing methods. In addition, we apply a VLM-based evaluation protocol to both simple and complex scenes, following recent approaches~\citep{hu2023tifa,li2024genai,yang2026hicogen}. Specifically, we evaluate prompt faithfulness along three dimensions: $\mathrm{Acc}_{\mathrm{exist}}$, $\mathrm{Acc}_{\mathrm{attribute}}$, and $\mathrm{Acc}_{\mathrm{relationship}}$, measuring element presence, attribute binding, and relationship correctness, respectively. Additionally, we report CLIPScore~\citep{hessel2021clipscore}, HPSv2~\citep{wu2023human}, and PickScore~\citep{kirstain2023pick} to assess global text--image alignment and perceptual quality. For consistency, we use Qwen3-VL-8B-Instruct~\citep{qwen3technicalreport} to construct layouts for all baselines and evaluate the corresponding prompt-faithfulness metrics. The VLM evaluation prompt templates are provided in Section~D of the supplementary material.

\begin{figure*}[!t]
\centering

\includegraphics[width=0.86\linewidth]{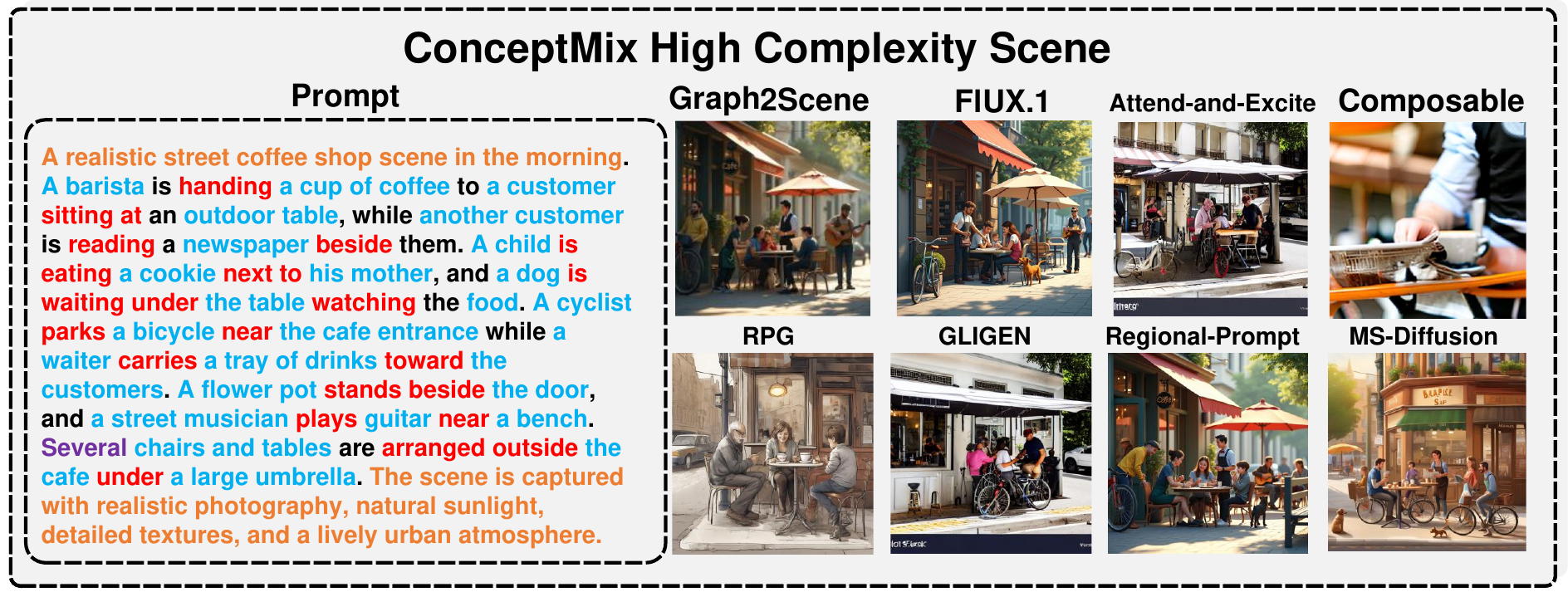}

\caption{
Visualization example of a complex scene in ConceptMix.
}

\label{fig:conceptmix_complexity}

\end{figure*}

\subsection{Comparison on Simple Scenes}
We first evaluate Graph2Scene on T2I-CompBench, which targets relatively simple scenes: object existence, attribute binding, and spatial composition. Existing methods improve these aspects by editing the prompt or adding spatial conditions. However, they mostly ground objects at the object level and leave the visual interactions to the diffusion model. When several objects compete for the same attention, attributes still leak between them and object counts still go wrong. Graph2Scene adds graph-guided self-attention, which lets related regions exchange information during denoising. This further improves quality, reaching 0.94, 0.80, and 0.89 on existence, attribute, and relationship accuracy, respectively. Compared with the strongest regional control baseline, Regional Prompt, Graph2Scene improves attribute accuracy from 0.69 to 0.80 and relationship accuracy from 0.73 to 0.89. This shows that explicit interaction modeling remains beneficial even for simple scenes.

\subsection{Comparison on Complex Scenes}

We then evaluate Graph2Scene on ConceptMix, where each prompt combines many concepts and interactions. As shown in Table~\ref{tab:conceptmix},  Graph2Scene ranks first on five of the six metrics and ties for the best on the last one. It reaches 0.74 on Acc$_{\text{attribute}}$ and 0.84 on  Acc$_{\text{relationship}}$. It also gives the best existence accuracy (0.88), and CLIPScore (26.45), and its HPSv2 (0.29) matches the top baseline. The gap is largest on relationships, which is exactly what our method targets.

Most baselines perform poorly on complex prompts, partly because several of them use older SD-based backbones, while layout-based methods mainly control where objects appear rather than how they interact. For a fairer comparison, we focus on methods using the same FLUX.1 backbone. The base FLUX.1 achieves 0.75 on Acc$_{\text{exist}}$ but only 0.42 on Acc$_{\text{relationship}}$. Regional Prompt improves attribute accuracy to 0.64 and relationship accuracy to 0.59 by assigning separate prompts to different regions, but many interactions are still missing. RPG shows a similar limitation on a weaker backbone, with only 0.09 relationship accuracy. These results indicate that region-level control helps object placement and attribute binding, but is still insufficient for modeling interactions in complex scenes.

Graph2Scene closes this gap. It does not stop at placing elements. Its graph edges let related regions share information during denoising. This way, the model builds the interaction between elements. On the same FLUX.1 backbone, the relationship score rises from 0.59 to 0.84, and attributes and existence also improve. This shows that the gain comes from relation-level control, not from a stronger base model. Additional qualitative comparisons on simple scenes and complex scenes at different complexity levels are provided in Section~G of the supplementary material.

\begin{figure*}[t]
    \centering

    \begin{subfigure}[t]{0.32\linewidth}
        \centering
        \includegraphics[width=\linewidth]{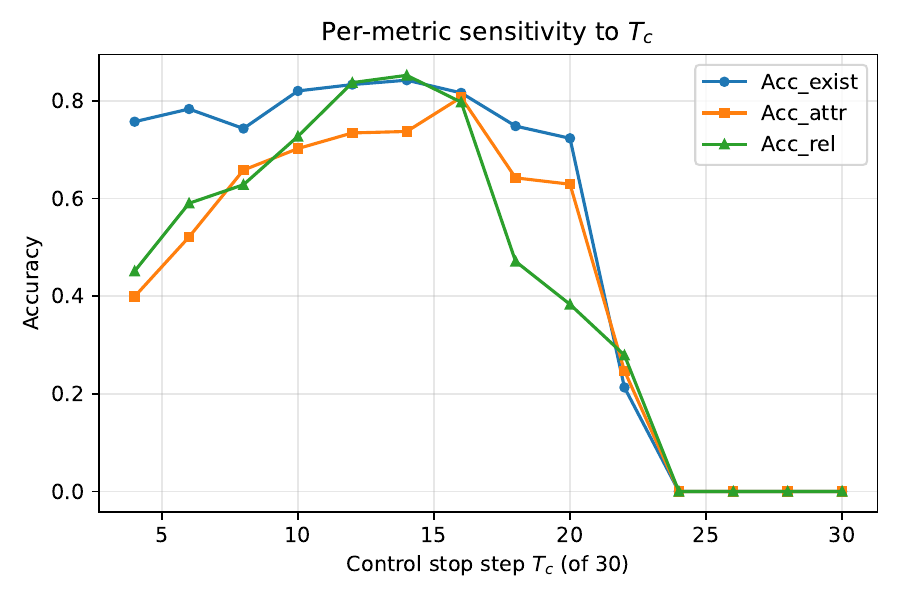}
        \caption{Per-metric sensitivity to control \mbox{horizon \(T_c\)}.}
        \label{fig:tc_permetric}
    \end{subfigure}
    \hfill
    \begin{subfigure}[t]{0.32\linewidth}
        \centering
        \includegraphics[width=\linewidth]{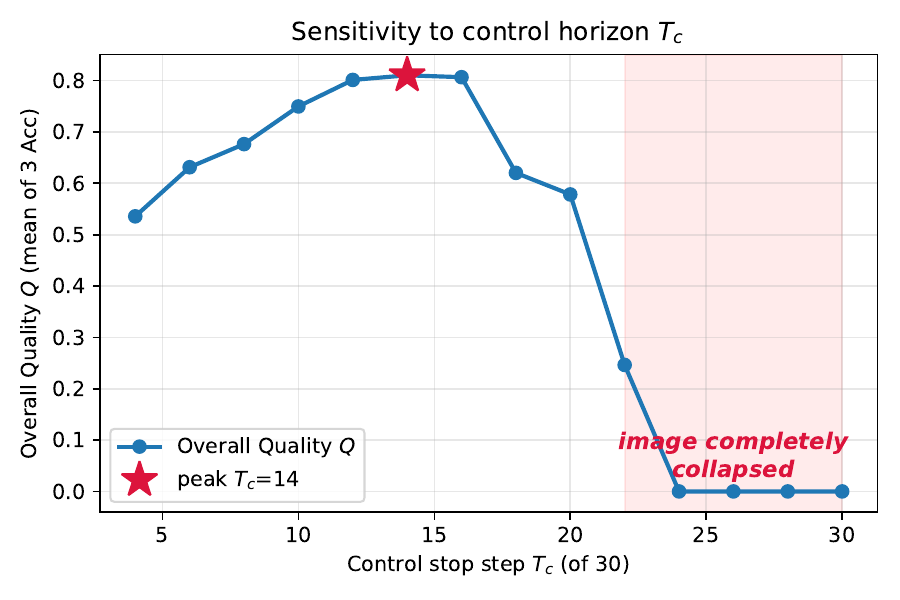}
        \caption{Overall-quality sensitivity to control \mbox{horizon \(T_c\)}.}
        \label{fig:tc_quality}
    \end{subfigure}
    \hfill
    \begin{subfigure}[t]{0.32\linewidth}
        \centering
        \includegraphics[width=\linewidth]{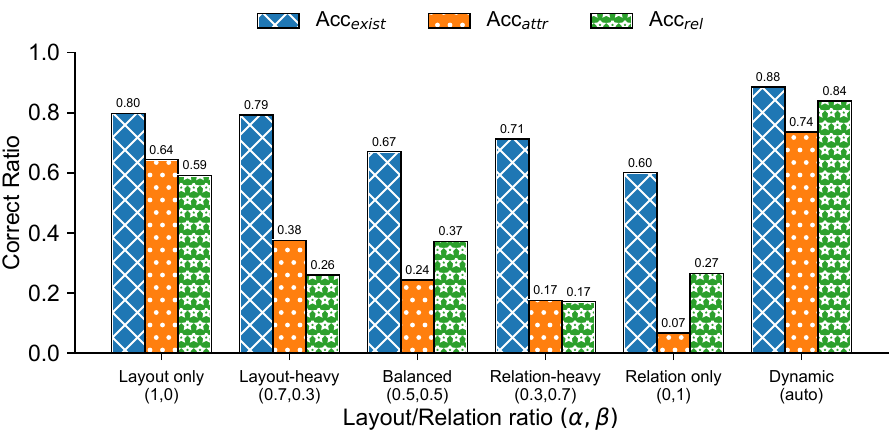}
        \caption{Sensitivity to the layout--relation control ratio.}
        \label{fig:ratio_mix}
    \end{subfigure}

    \caption{Parameter sensitivity analysis of Graph2Scene.}
    \label{fig:control_ablation}

\end{figure*}

\subsection{Parameter Sensitivity Analysis}
\paragraph{Effect of the control horizon $T_c$.}
The control horizon $T_c$ determines how many denoising steps apply graph control. Figure~\ref{fig:control_ablation} shows the effect of different control horizons on both individual metrics and overall quality. As shown in Figure~\ref{fig:tc_quality}, the overall quality improves from 0.54 at $T_c{=}4$ to the best score of 0.81 at $T_c{=}14$, but decreases when the control is applied for too long. Figure~\ref{fig:tc_permetric} further shows that object existence improves quickly with short control (0.76 at $T_c{=}4$), while attributes and relationships require longer guidance to form. After $T_c{=}14$, excessive control starts to harm generation. The quality drops to 0.25 at $T_c{=}22$, and the images collapse when $T_c{\ge}24$.
This is because later denoising steps mainly refine visual details, where strong control can interfere with the model's natural generation process. See Supplementary Section~B.

\smallskip
\noindent\textbf{Effect of the layout--relation ratio $(\alpha,\beta)$.}
We next fix the two control strengths to different values and compare them with our dynamic schedule. As shown in Figure~\ref{fig:ratio_mix}, no fixed ratio works well. Using layout alone $(1,0)$ already gives 0.64 on attributes and 0.59 on relationships. Surprisingly, any fixed mix does worse. The balanced setting $(0.5,0.5)$ drops to 0.24 and 0.37, and the relation-only setting $(0,1)$ drops to 0.07 and 0.27. The reason is timing. A fixed $\beta$ applies relation control from the very first step, before the layout is settled. The two controls then fight each other, and both attributes and relationships suffer. Our dynamic schedule avoids this. It lets layout form first, adds relation control later, and then releases both. This raises all three metrics to the highest values (0.88, 0.74, 0.84), and matches how diffusion models build an image from layout to detail.

\begin{table}[!t]
\centering
\caption{Ablation study of Graph2Scene components and control strategies.
L+R denotes Layout and Relation Guidance; Dyn. denotes dynamic ratio adjustment.}
\label{tab:ablation}
\small
\setlength{\tabcolsep}{3pt}
\renewcommand{\arraystretch}{1.04}
\begin{tabular}{lccc}
\toprule
Configuration
& Acc$_{\text{exist}}$ $\uparrow$
& Acc$_{\text{attr}}$ $\uparrow$
& Acc$_{\text{rel}}$ $\uparrow$ \\
\midrule
FLUX.1 (Original) & 0.75 & 0.31 & 0.42 \\
FLUX.1 (Rewritten) & 0.69 & 0.32 & 0.35 \\
\midrule
Layout-only ($T_c=15$) & 0.86 & 0.61 & 0.53 \\
Layout-only ($T_c=20$) & 0.79 & 0.53 & 0.52 \\
Layout-only (Adaptive $T_c$) & 0.80 & 0.64 & 0.59 \\
Relation-only ($T_c=15$) & 0.55 & 0.15 & 0.37 \\
Relation-only ($T_c=20$) & 0.66 & 0.09 & 0.24 \\
Relation-only (Adaptive $T_c$) & 0.60 & 0.07 & 0.27 \\
\midrule
L+R (Dyn., $T_c=15$) & 0.80 & 0.71 & 0.83 \\
L+R (Dyn., $T_c=20$) & 0.72 & 0.64 & 0.38 \\
\textbf{L+R (Dyn., Adaptive $T_c$)}
& \textbf{0.88} & \textbf{0.74} & \textbf{0.84} \\
\bottomrule
\end{tabular}
\end{table}

\subsection{Ablation Study}
\smallskip
\noindent\textbf{Prompt Rewriting.}
Prompt rewriting alone reduces existence and relationship accuracy from 0.75 to 0.69 and from 0.42 to 0.35, respectively, indicating that structured scene descriptions alone are insufficient to improve generation.

\smallskip
\Needspace{3\baselineskip}
\noindent\textbf{Effectiveness of Layout and Relation Guidance.}
As shown in Table~\ref{tab:ablation}, layout-only guidance improves attribute accuracy from 0.31 to 0.64. In contrast, relation-only guidance drops it to 0.07, indicating that relation guidance alone is insufficient to preserve attribute correctness. Combining both mechanisms achieves higher attribute and relationship accuracy of 0.74 and 0.84, respectively, demonstrating their complementary roles in improving generation quality.

\smallskip
\noindent\textbf{Effect of Adaptive Control Step.}
We further compare fixed and adaptive control durations. With both guidance mechanisms enabled, adaptive control achieves the highest mean accuracy across the three metrics among the evaluated settings, reaching 0.88, 0.74, and 0.84 in existence, attribute, and relationship accuracy, respectively. Compared with fixed control at $T_c=15$, it improves existence and relationship accuracy from 0.80 to 0.88 and from 0.83 to 0.84, respectively, while maintaining higher attribute accuracy (0.74 vs. 0.71). These results suggest that adaptive control better balances object placement, attribute preservation, and relationship generation.

\section{Conclusion}

In this paper, we present Graph2Scene, a training-free framework that compiles scene graphs into attention control for compositional text-to-image generation. Based on the observation that existing attention-control methods mainly focus on individual element placement while overlooking inter-object relations, Graph2Scene explicitly models both elements and their interactions through graph-guided attention. Our framework converts scene graph structures into layout- and relation-aware attention constraints, and applies them at different denoising stages with a complexity-aware control schedule.
Graph2Scene is compatible with existing diffusion models without additional training and can be directly applied to complex prompts with multiple interacting elements. Extensive experiments demonstrate that Graph2Scene improves compositional generation quality, especially for complex scenes with dense interactions.

\label{main:body-end}
\clearpage
\section{LLM Usage Statement}
\label{main:disclosure}
In this work, we used generative AI tools (large language models) to proofread the paper, edit it for readability, and review the manuscript before submission. In addition, our experiments use the ConceptMix benchmark, which requires specifying the complexity level $K$ and constructing prompts with different numbers of compositional concepts, following the evaluation protocol of the original work~\citep{wu2024conceptmix}. Generative AI was used to assist in constructing some of these benchmark prompts under the predefined $K$ settings and evaluation rules. All generated prompts were manually reviewed to ensure that they satisfied the benchmark requirements. We take responsibility for the final content of this work, including the manuscript text and experimental prompts produced with the assistance of generative AI.

{
\small
\bibliographystyle{ieeenat_fullname}
\bibliography{references}
}

\input{supplementary}
\end{document}

%% file: math_commands.tex
\usepackage{amsmath,amsfonts,bm}

\def\eqref#1{equation~\ref{#1}}

\def\1{\bm{1}}

\DeclareMathAlphabet{\mathsfit}{\encodingdefault}{\sfdefault}{m}{sl}
\SetMathAlphabet{\mathsfit}{bold}{\encodingdefault}{\sfdefault}{bx}{n}



%% file: supplementary.tex

\onecolumn
\makeatletter
\renewcommand\normalsize{%
  \@setfontsize\normalsize{10.95pt}{13.6pt}%
  \abovedisplayskip 11pt plus3pt minus6pt
  \abovedisplayshortskip 0pt plus3pt
  \belowdisplayshortskip 6.5pt plus3.5pt minus3pt
  \belowdisplayskip\abovedisplayskip
  \let\@listi\@listI}
\renewcommand\small{%
  \@setfontsize\small{10pt}{12pt}%
  \abovedisplayskip 10pt plus2pt minus5pt
  \abovedisplayshortskip 0pt plus3pt
  \belowdisplayshortskip 6pt plus3pt minus3pt
  \belowdisplayskip\abovedisplayskip}
\renewcommand\footnotesize{\@setfontsize\footnotesize{9pt}{11pt}}
\renewcommand\scriptsize{\@setfontsize\scriptsize{8pt}{9.5pt}}
\renewcommand\tiny{\@setfontsize\tiny{6pt}{7pt}}
\setlength{\@fpsep}{12pt}
\makeatother
\normalsize
\setlength{\topskip}{11pt}
\setlength{\maxdepth}{5.5pt}
\raggedbottom
\renewcommand{\topfraction}{0.95}
\renewcommand{\bottomfraction}{0.90}
\renewcommand{\textfraction}{0.05}
\renewcommand{\floatpagefraction}{0.67}
\setcounter{totalnumber}{3}
\setcounter{topnumber}{2}
\setcounter{bottomnumber}{2}
\setlength{\textfloatsep}{12pt plus3pt minus3pt}
\setlength{\intextsep}{12pt plus3pt minus3pt}
\setlength{\floatsep}{10pt plus3pt minus2pt}
\captionsetup{skip=5pt}
\setcounter{figure}{0}
\setcounter{table}{0}
\setcounter{equation}{0}
\renewcommand{\theHfigure}{supp.\arabic{figure}}
\renewcommand{\theHtable}{supp.\arabic{table}}
\renewcommand{\theHequation}{supp.\arabic{equation}}

\begin{center}
  {\LARGE\bfseries\thetitle\par}
  \vspace{0.4em}
  {\large Supplementary Material\par}
\end{center}

This supplement provides implementation details, additional experiments, and qualitative comparisons for Graph2Scene.

\noindent
\begin{minipage}{\linewidth}
\captionsetup{type=table,skip=5pt}
\caption*{\textbf{Overview of Supplementary Material}}
\small
\setlength{\tabcolsep}{5pt}
\renewcommand{\arraystretch}{1.05}
\begin{tabularx}{\linewidth}
{@{}>{\raggedright\arraybackslash}p{0.35\linewidth}
>{\raggedright\arraybackslash}X@{}}
\toprule
\textbf{Section} & \textbf{Summary} \\
\midrule

\SectionEntry{supp:app:graph_construction_details}
{Graph Construction Details}
& Describes how a VLM organizes scene information, elements with bounding boxes, and relations into a scene graph and a structured prompt. \\

\SectionEntry{supp:app:control_duration}
{Analysis of Graph Control Duration}
& Examines the artifacts caused by prolonged graph guidance and motivates the choice of a maximum control horizon. \\

\SectionEntry{supp:sec:sg_comparison}
{Additional Comparison with Scene Graph-Based Methods}
& Compares Graph2Scene with scene graph baselines on COCO-Stuff and examines their outputs on complex open prompts. \\

\SectionEntry{supp:app:prompts}
{Prompt and Evaluation Templates}
& Provides the scene graph construction prompt and VLM prompts for evaluating object existence, attribute correctness, and relationship accuracy. \\

\SectionEntry{supp:app:Details}
{Implementation Details of the Graph Execution Scheduler}
& Defines attention leakage and overlap and shows how they determine the adaptive control duration. \\

\SectionEntry{supp:app:runtime_overhead}
{Runtime Overhead Analysis}
& Compares stage-level runtime with Regional Prompt across prompt lengths and image resolutions and reports compositional quality for prompt scaling. \\

\SectionEntry{supp:app:longbench}
{Additional Qualitative Results}
& Shows qualitative comparisons on simple scenes, ConceptMix examples at three complexity levels, and long prompts from LongBench-T2I. \\

\bottomrule
\end{tabularx}
\end{minipage}
\par\medskip
\appendix
\renewcommand{\theHsection}{supp.\Alph{section}}

\section{Graph Construction Details}
\label{supp:app:graph_construction_details}

\begin{figure}[!htbp]
    \centering
    \includegraphics[
        width=0.90\linewidth,
        height=0.75\textheight,
        keepaspectratio
    ]{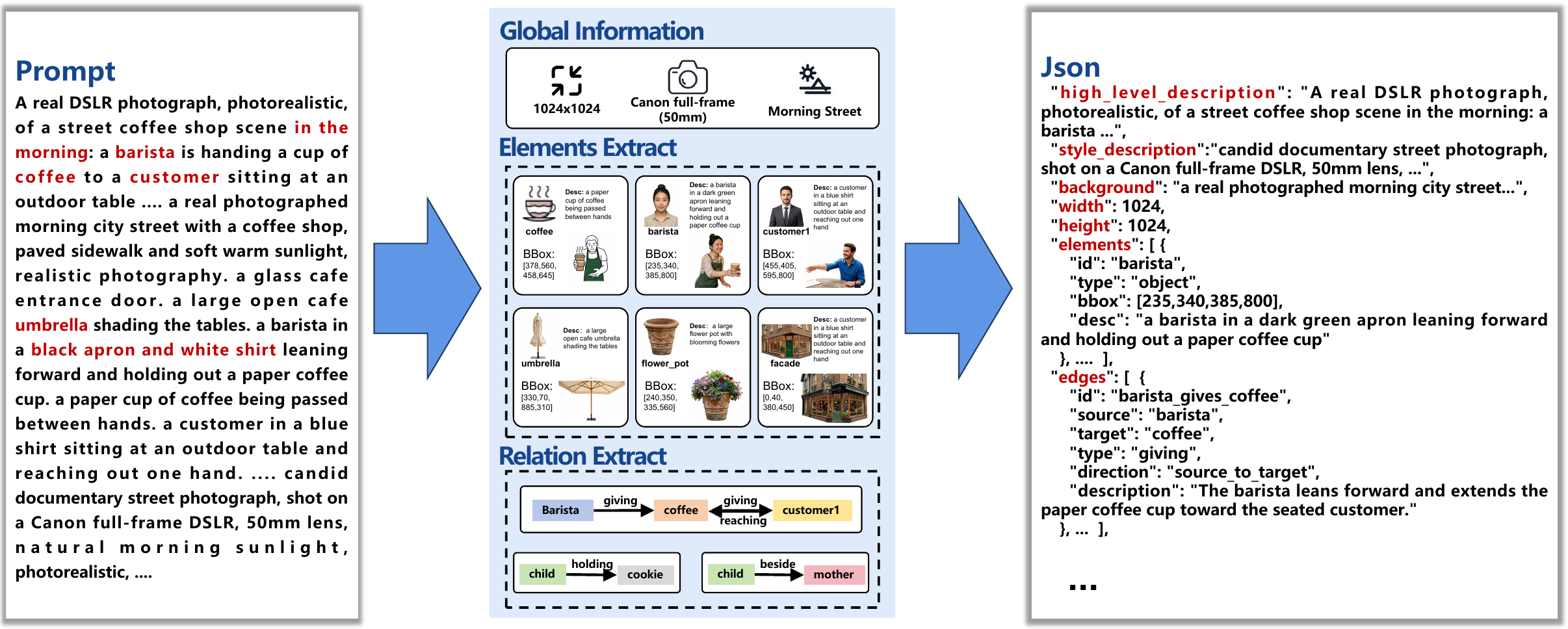}
    \caption{Prompt-to-JSON conversion template used in Graph2Scene.}
    \label{supp:fig:json}
\end{figure}

\paragraph{Graph construction.}
Graph2Scene constructs the scene graph directly from the input prompt using Qwen3-VL-8B-Instruct~\citep{qwen3technicalreport}. As illustrated in Figure~\ref{supp:fig:json}, we prompt the VLM to organize the scene information into three parts: global information describing the overall scene and style, individual elements with their descriptions and predicted bounding boxes, and relations specifying how pairs of elements interact. The extracted information is returned in a structured JSON format, where elements correspond to graph nodes and relations correspond to graph edges.

\paragraph{Node properties.}
Each element node contains four basic properties. \texttt{id} uniquely identifies the element, while \texttt{type} records its semantic category. \texttt{bbox} specifies the predicted spatial region of the element in the image, and \texttt{desc} provides its local visual description, including appearance and attributes. These properties provide the semantic and spatial information required to associate each element with its corresponding text and image tokens.

\paragraph{Edge properties.}
Each relation edge describes an interaction between two element nodes. \texttt{source} and \texttt{target} specify the participating elements, \texttt{type} records the relation category, and \texttt{direction} indicates its direction when applicable. \texttt{description} provides a natural-language description of the relation, which is used to identify the corresponding relation tokens for subsequent attention control. Each edge also has a unique \texttt{id} for reference.

\paragraph{Style and global organization.}
In addition to elements and relations, we separately organize scene-level information that applies to the entire image. This information includes the background, overall visual style, lighting, viewpoint, and other imaging conditions. Such global descriptions are shared across all elements and relations rather than being associated with a specific node or edge.

\Needspace{5\baselineskip}
\begin{figure}[!htbp]
    \centering
    \includegraphics[width=0.70\linewidth]{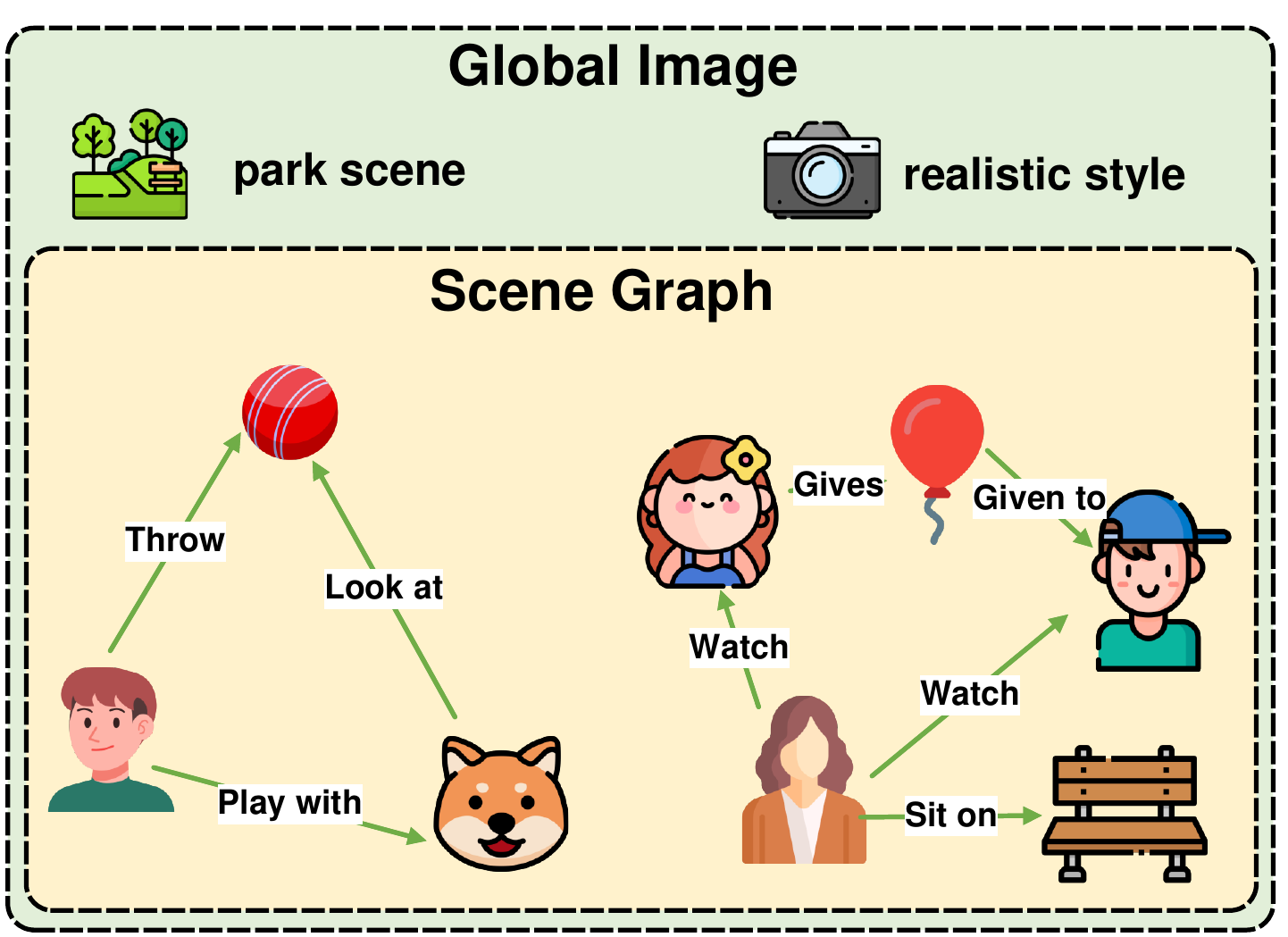}
    \caption{Organization of image information in Graph2Scene.}
    \label{supp:fig:graph}
\end{figure}
\paragraph{Prompt organization.}
During graph construction, the VLM also organizes the original prompt according to the extracted structure. Scene-level descriptions are retained as global information, while element and relation descriptions are associated with their corresponding nodes and edges. The resulting structured prompt is used as the textual condition for generation, allowing Graph2Scene to identify the text tokens associated with each element and relation during attention control.

\section{Analysis of Graph Control Duration}
\label{supp:app:control_duration}

Modern diffusion Transformers are pretrained with full attention, where each token can interact with all other tokens during denoising. Graph2Scene modifies this process at inference time by restricting some interactions according to the scene graph. Such control helps the model better establish object layouts and relations. However, the pretrained model is not optimized for this modified attention pattern throughout the entire denoising process. If the control is applied for too many steps, the model increasingly deviates from its original full-attention behavior and cannot properly adapt to the modified attention computation~\cite{chen2024training,wu2023harnessing}. As a result, the generation process becomes unstable and may eventually produce severe visual artifacts.

\begin{figure}[!htbp]
    \centering
    \includegraphics[
        width=0.95\linewidth,
        keepaspectratio
    ]{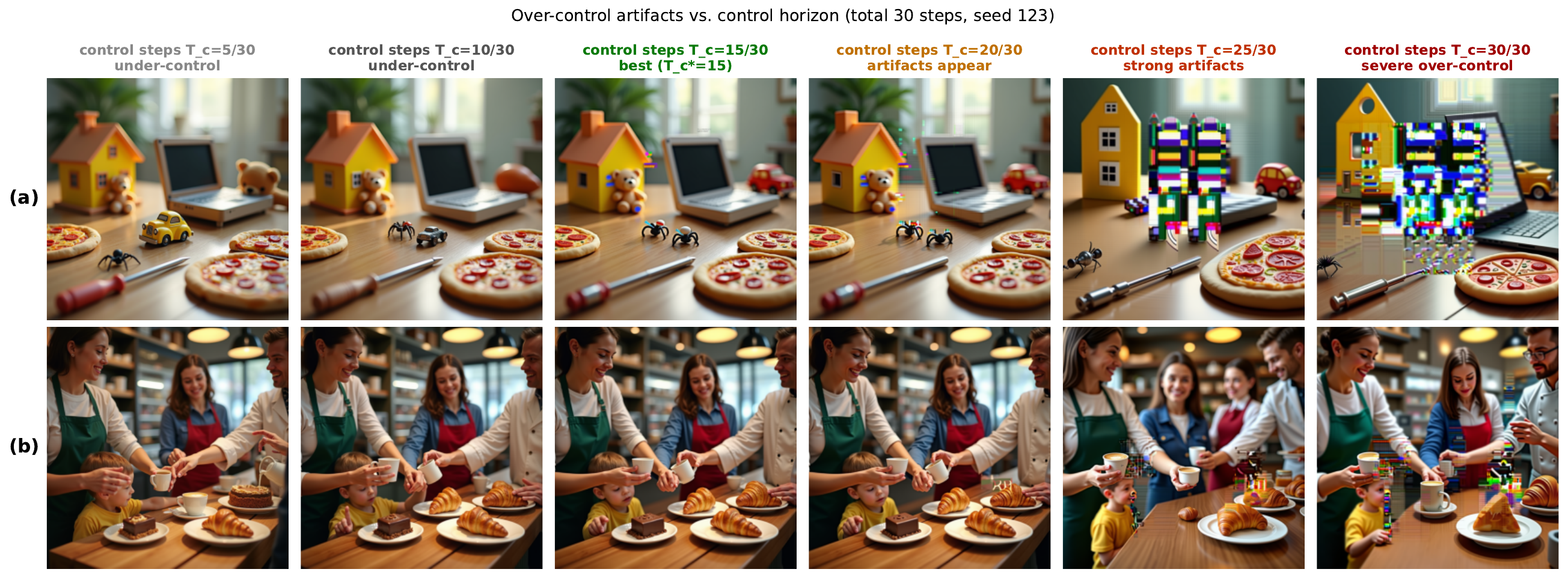}
    \caption{Examples of artifacts caused by excessive control.}
    \label{supp:fig:artifacts}
\end{figure}

This phenomenon is also consistently observed in our experiments. As shown quantitatively in Figure~6(a) and~(b) of the main paper,
generation quality drops sharply when the control is extended beyond about 20 denoising steps. When the control horizon reaches 25 steps or more, many generated images are close to complete collapse and no longer contain recognizable visual content. The qualitative examples in
Figure~\ref{supp:fig:artifacts} further illustrate this behavior. Slight artifacts already appear around step 20, and become increasingly severe as the control duration increases. These results show that, under the attention modification used by Graph2Scene, longer control does not continuously improve generation. Instead, excessive control increasingly interferes with the pretrained model's original full-attention generation process and eventually leads to unstable generation or visual collapse.

Based on these observations and experimental results, we set the maximum control horizon to $T_c = 0.6T$.

\Needspace{6\baselineskip}
\section{Additional Comparison with Scene Graph-Based Methods}
\label{supp:sec:sg_comparison}
Scene graph-based generation methods aim to address the same fundamental challenge as Graph2Scene: generating images with accurate objects, attributes, and relational interactions from structured descriptions. Therefore, we further compare Graph2Scene with representative scene graph-based approaches, including \textbf{SG2IM}~\cite{johnson2018image}, \textbf{SGDiff}~\cite{yang2022diffusion}, and \textbf{CanonicalSg2Im}~\cite{herzig2020learning}, to evaluate the effectiveness of explicitly executing scene graph constraints during the generation process.
\begin{table}[!htbp]
\centering
\caption{Quantitative comparison with scene graph-based methods.}
\label{supp:tab:scene_graph_comparison}
\small
\renewcommand{\arraystretch}{1.16}
\begin{tabular*}{\linewidth}{@{\extracolsep{\fill}}lcccccc@{}}
\toprule
Method & Acc$_{\text{exist}}$ $\uparrow$ & Acc$_{\text{attribute}}$ $\uparrow$ & Acc$_{\text{relationship}}$ $\uparrow$ & CLIPScore $\uparrow$ & HPSv2 $\uparrow$ & PickScore $\uparrow$ \\
\midrule
SG2IM & 0.553 & 0.286 & 0.366 & 16.552 & 0.125 & 17.976 \\
SGDiff & 0.336 & 0.208 & 0.253 & 19.307 & 0.131 & 18.453 \\
CanonicalSg2Im & 0.032 & 0.016 & 0.000 & 12.195 & 0.085 & 16.349 \\
\midrule
Graph2Scene & \textbf{0.927} & \textbf{0.893} & \textbf{0.872} & \textbf{26.296} & \textbf{0.294} & \textbf{22.845} \\
\bottomrule
\end{tabular*}
\end{table}

\paragraph{Experiment Setting.}
All compared scene graph-based methods provide publicly available models trained on the COCO dataset~\cite{lin2014microsoft}.
We directly use their released COCO checkpoints for inference and evaluate all methods on the COCO-Stuff dataset~\cite{caesar2018coco} with an NVIDIA L20 GPU under the same experimental setting.
Following our main experiments, we adopt the same evaluation metrics, including object existence accuracy ($\mathrm{Acc}_{\mathrm{exist}}$), attribute accuracy ($\mathrm{Acc}_{\mathrm{attribute}}$), relationship accuracy ($\mathrm{Acc}_{\mathrm{relationship}}$), CLIPScore, HPSv2, and PickScore, to measure both semantic alignment and overall generation quality.

\paragraph{Quantitative Comparison.}
Table~\ref{supp:tab:scene_graph_comparison} shows that Graph2Scene consistently outperforms existing scene graph-based methods across all metrics. SG2IM uses graph convolutions to predict layouts and a GAN-based refinement network for generation.
SGDiff employs masked contrastive pre-training to encode scene graphs into a latent diffusion model. CanonicalSg2Im learns canonical graph representations to handle semantically equivalent graphs. However, these methods struggle with visual fidelity due to limited generation capacity and imperfect alignment between graph structures and visual content. In contrast, Graph2Scene leverages a powerful pretrained text-to-image diffusion backbone and explicitly compiles scene graph structures into attention operations. This design integrates structural constraints while preserving the model's native generation ability.
As a result, Graph2Scene achieves substantial improvements in object existence, attribute recognition, and relationship accuracy, reaching scores of 0.927, 0.893, and 0.872, respectively. These results demonstrate more effective preservation of scene semantics and interactions compared with previous scene graph-based approaches. Moreover, Graph2Scene obtains higher overall visual quality scores, achieving a CLIPScore of 26.296, an HPSv2 score of 0.294, and a PickScore of 22.845, indicating better alignment with both image-text similarity and human preference.

\begin{figure}[!htbp]
    \centering
    \includegraphics[
        width=\linewidth,
        keepaspectratio
    ]{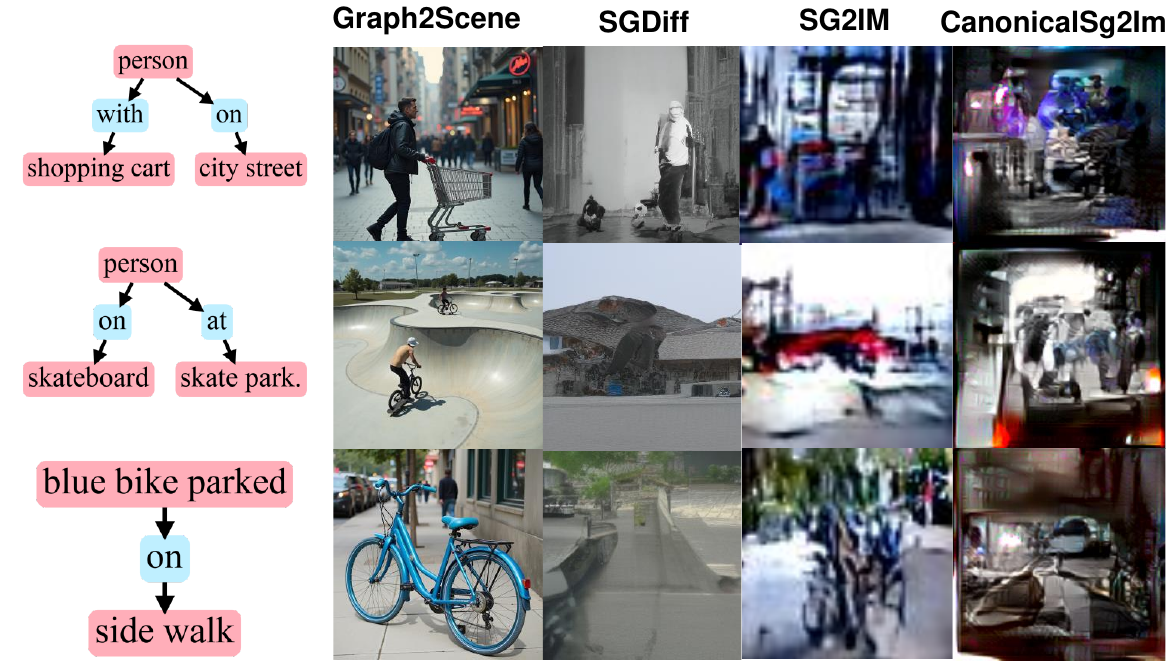}
    \caption{Case studies comparing Graph2Scene with existing Scene Graph based methods on the COCO-Stuff dataset.}
    \label{supp:fig:coco}
\end{figure}

\paragraph{Case Study on COCO-Stuff.}
Figure~\ref{supp:fig:coco} presents qualitative comparisons on the COCO-Stuff dataset.
Although all methods receive the same scene graph conditions, Graph2Scene produces significantly clearer and more realistic images while preserving the specified objects and relationships.
This improvement is mainly attributed to the stronger generative capability of modern diffusion backbones and the explicit execution of scene graph constraints during generation.
In contrast, previous scene graph-based methods are built upon earlier generation architectures, such as GAN-based pipelines or early scene graph-conditioned diffusion models, which often suffer from limited visual fidelity and difficulty in aligning graph structures with image contents.
As shown in Figure~\ref{supp:fig:coco}, SGDiff can generate some recognizable scene elements, such as the person and skateboard, but many details remain blurry and incomplete.
For SG2IM and CanonicalSg2Im, although the generated images contain rough color distributions and coarse object regions, the objects are often heavily distorted, making their categories and attributes difficult to identify.
For example, the shopping cart, city street, and bicycle structures are only weakly represented, and the generated contents lack clear boundaries and realistic appearances.

\begin{figure}[!htbp]
    \centering
    \includegraphics[
        width=\linewidth,
        keepaspectratio
    ]{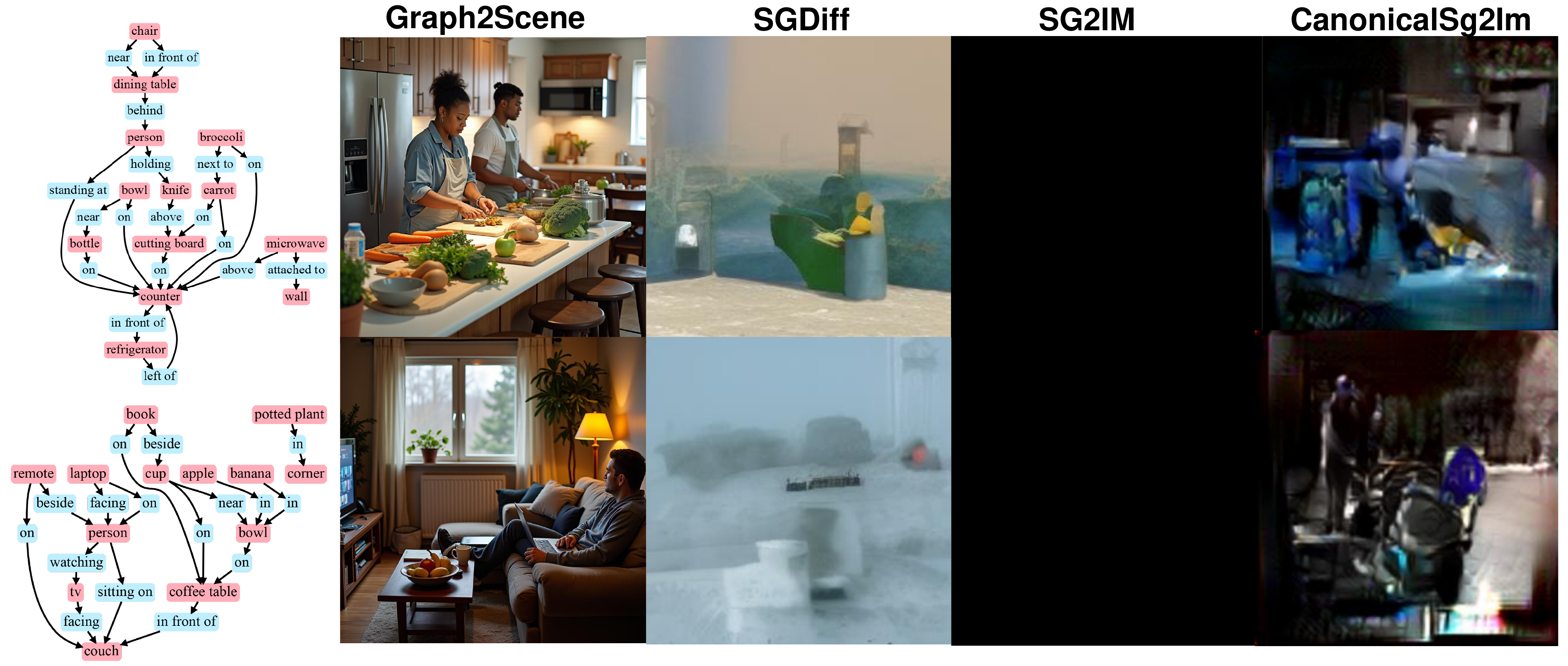}
    \caption{Case studies comparing Graph2Scene with existing scene graph-based methods on the open complex prompts.}
    \label{supp:fig:open}
\end{figure}

\Needspace{4\baselineskip}
\paragraph{Case Study on Complex Open Prompts.}
Figure~\ref{supp:fig:open} further presents qualitative comparisons on open complex prompts with larger and more diverse scene graphs.
Compared with the COCO-Stuff examples, these prompts contain more objects, attributes, and relational constraints, posing greater challenges for scene graph-based generation.
Graph2Scene consistently generates clear and coherent images while preserving multiple objects and their interactions, demonstrating its ability to generalize to complex scene structures.
In contrast, previous methods suffer from severe degradation when handling such complex inputs.
SGDiff can still generate some rough scene layouts, but the generated images remain highly blurred and fail to recover detailed object appearances.
For SG2IM, the model completely collapses and produces nearly black images under these open complex prompts.
This failure is mainly because SG2IM relies on a graph encoder and layout prediction pipeline learned from fixed training distributions, making it difficult to handle unseen graph compositions and substantially larger scene structures.

\Needspace{0.95\textheight}
\section{Prompt and Evaluation Templates}
\label{supp:app:prompts}
This section provides the prompt templates used in Graph2Scene. We include the VLM prompt used to construct the structured scene graph from the input text, together with the templates used for VLM-based evaluation of object existence, attribute correctness, and relationship accuracy. The complete prompt formats are shown in the following figures.

\begin{figure}[H]
    \centering
    \includegraphics[
        width=0.98\linewidth,
        height=0.80\textheight,
        keepaspectratio
    ]{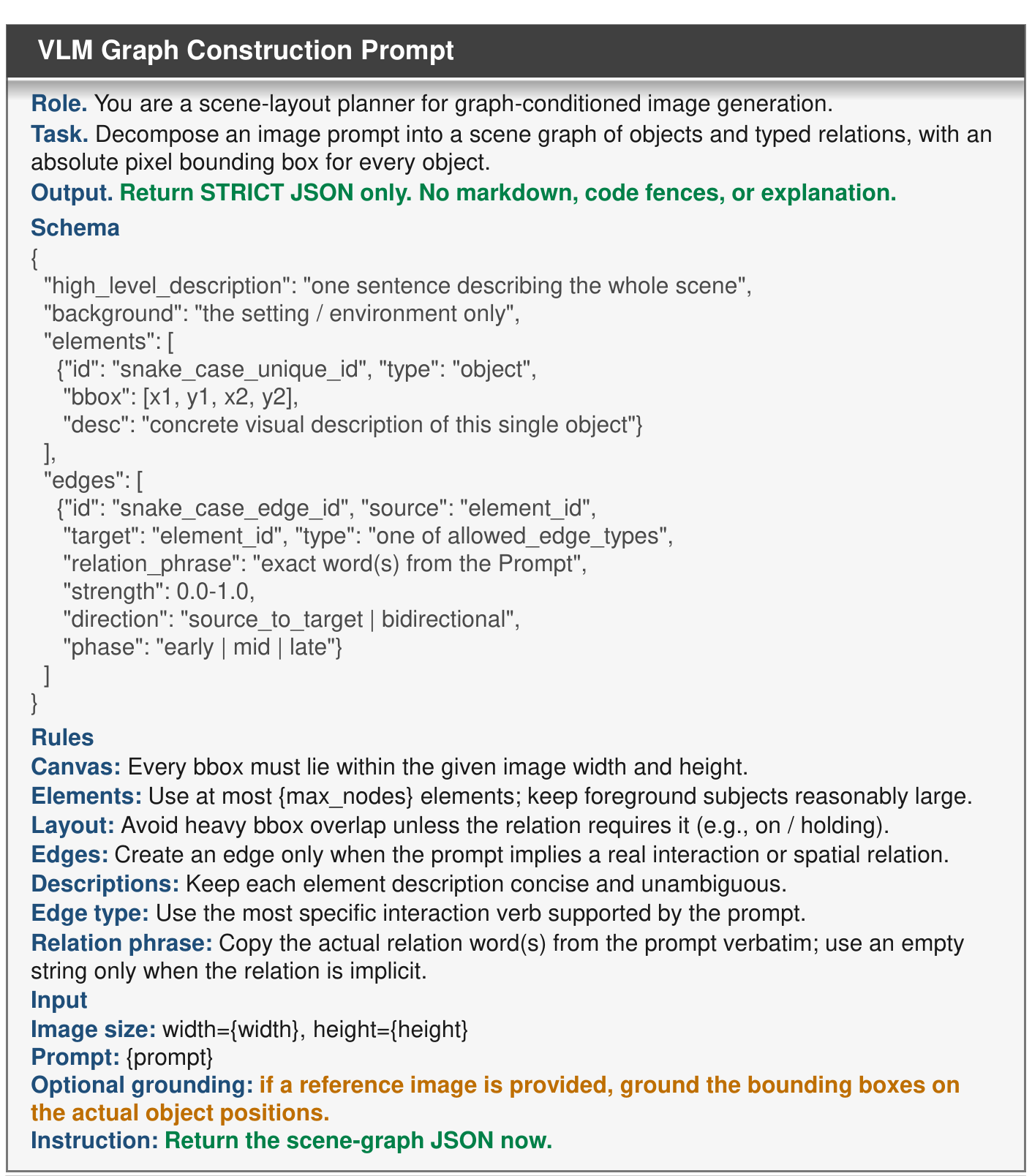}
    \caption{Prompt template used for scene graph construction.}
    \label{supp:fig:graph_construction_prompt}
\end{figure}

\begin{figure}[!htbp]
\centering

\par\noindent\includegraphics[width=0.98\linewidth]{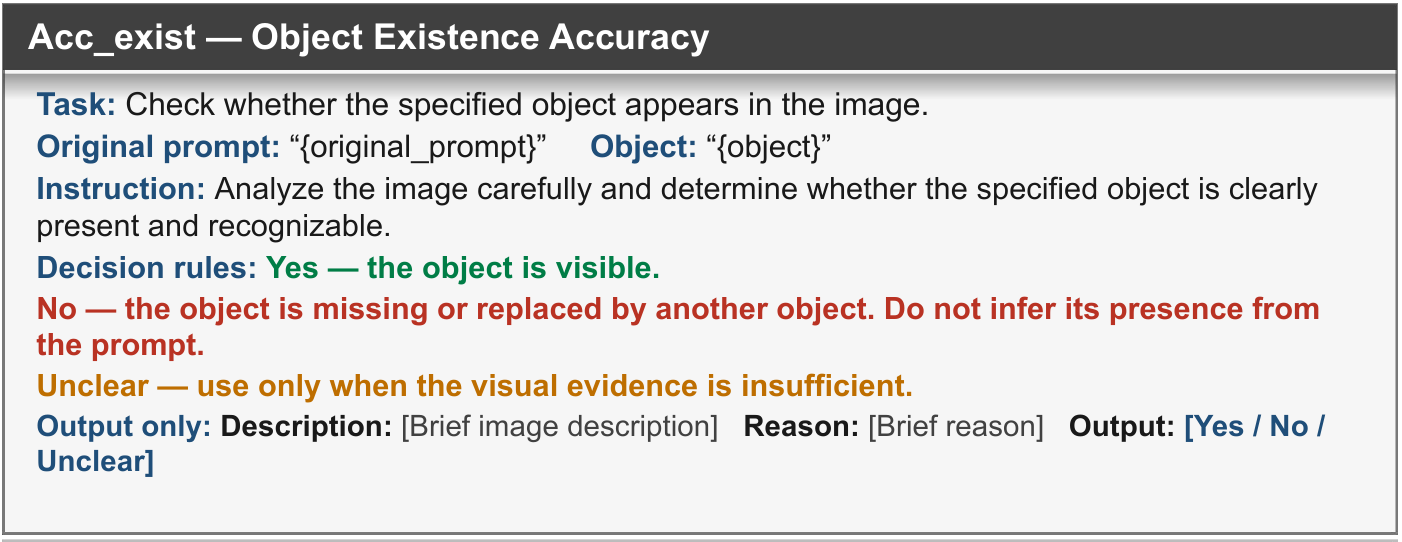}\par
\smallskip
\par\noindent\includegraphics[width=0.98\linewidth]{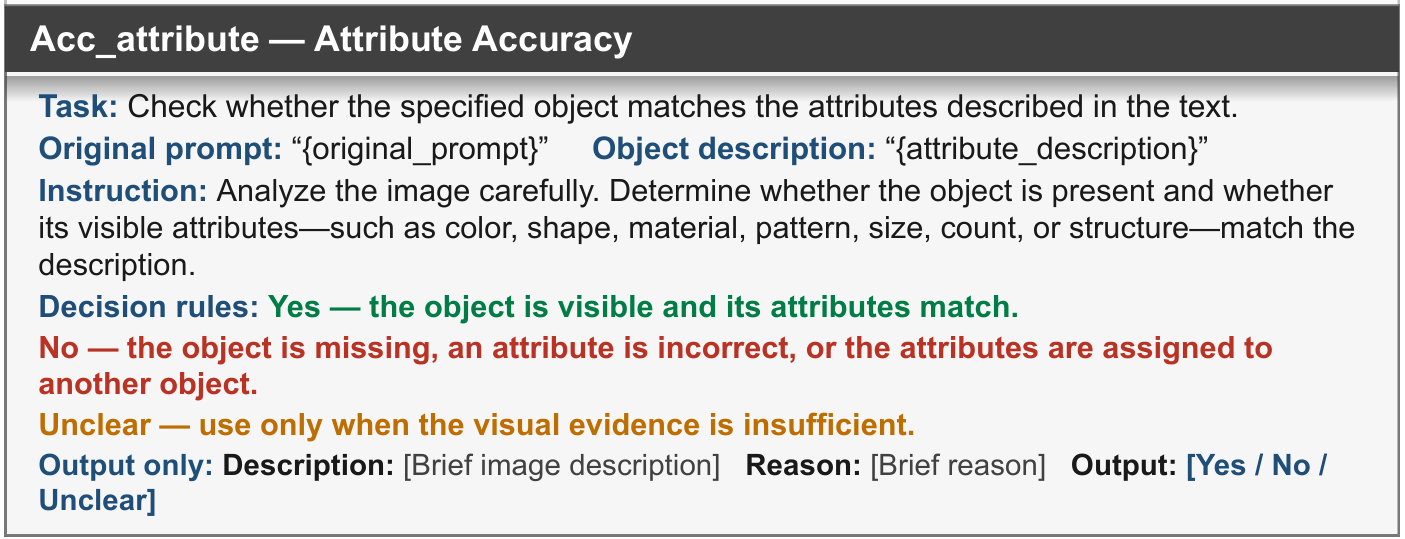}\par
\smallskip
\par\noindent\includegraphics[width=0.98\linewidth]{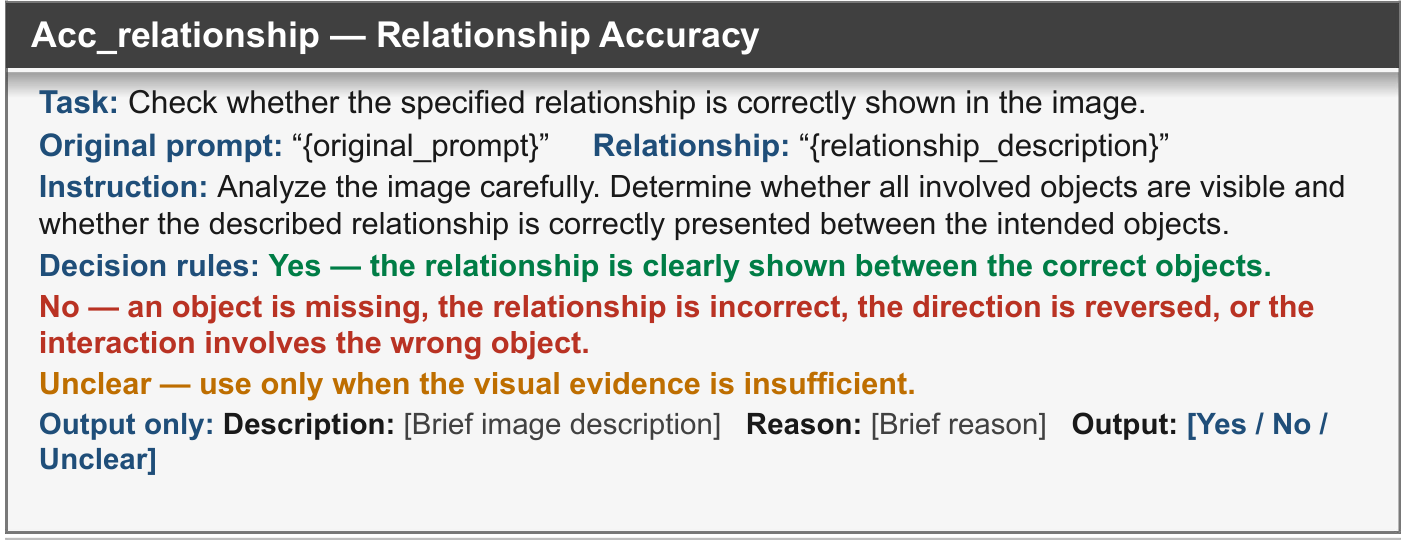}\par

\caption{Prompt templates used to evaluate
$\mathrm{Acc}_{\mathrm{exist}}$,
$\mathrm{Acc}_{\mathrm{attribute}}$, and
$\mathrm{Acc}_{\mathrm{relationship}}$
with a VLM.}
\label{supp:fig:evaluation_templates}
\end{figure}

\FloatBarrier
\section{Implementation Details of the Graph Execution Scheduler}
\label{supp:app:Details}
 We extract element tokens and relation tokens from the prompt, denoted by $\mathcal{U}_{\mathrm{elem}}$ and $\mathcal{U}_{\mathrm{rel}}$, respectively. For each token $u$, we collect its cross-attention maps over all denoising steps and aggregate them into a~single~attention~map~$A_u$.

For each token $u$, we obtain a set of high-response regions and their normalized attention scores, denoted by
\begin{equation}
\mathcal{C}_u=\{c_u^1,c_u^2,\ldots\},
\end{equation}
where each $c\in\mathcal{C}_u$ denotes the attention score of a high-response region.

\paragraph{Attention leakage.}
We regard the region with the highest attention score as the dominant response of token $u$, and treat the remaining response as attention leakage:
\begin{equation}
L_u
=
1-\max_{c\in\mathcal{C}_u} c.
\end{equation}

\paragraph{Attention overlap.}
Attention overlap measures the spatial confusion between concepts of the same type. We compute overlap only between element tokens or between relation tokens, since relation attention can naturally overlap with its participating elements. Let $\mathcal{H}_u$ denote the high-response spatial region of token $u$. For $u\neq v$, where either $u,v\in\mathcal{U}_{\mathrm{elem}}$ or $u,v\in\mathcal{U}_{\mathrm{rel}}$, we define
\begin{equation}
O_{u,v}
=
\frac{
|\mathcal{H}_u \cap \mathcal{H}_v|
}{
|\mathcal{H}_u \cup \mathcal{H}_v|
}.
\end{equation}
After computing the two attention statistics, we take the largest leakage value over all element and relation tokens as $L_{\max}$, and the largest overlap value among element-element and relation-relation token pairs as $O_{\max}$.

We use the more severe attention failure to determine the control ratio and the corresponding control duration:
\begin{equation}
\rho=\max(L_{\max},O_{\max}),
\qquad
T_{\mathrm{ctrl}}=\rho T_c,
\end{equation}
where $\rho\in[0,1]$ represents the required control level and $T_c$ denotes the maximum control duration, whose setting is discussed in Appendix~\ref{supp:app:control_duration}.

\section{Runtime Overhead Analysis}
\label{supp:app:runtime_overhead}

We conduct runtime experiments on a single NVIDIA L20 GPU to evaluate the computational overhead of Graph2Scene. The VLM planning stage uses Qwen3-VL-8B-Instruct~\citep{qwen3technicalreport}, and all evaluation prompts are drawn from the ConceptMix benchmark~\citep{wu2024conceptmix} introduced in the main paper. All runtime measurements are averaged over five runs.

\paragraph{Experimental Setup.}
We investigate runtime overhead from two perspectives. First, we evaluate prompts with different lengths (200--400, 400--600, 600--800, and 800--1000 tokens) while keeping the image resolution fixed at $1024\times1024$. This experiment examines how prompt complexity affects both computational cost and compositional quality. Second, we fix the prompt length to 400--600 tokens and vary the image resolution from $512\times512$ to $1280\times1280$ to investigate resolution-dependent computational overhead. Since this second experiment focuses on runtime scaling, we report only runtime measurements in Table~\ref{supp:tab:runtime_breakdown}(b). VLM-based quality metrics are evaluated and reported in the prompt scaling experiment in Table~\ref{supp:tab:runtime_breakdown}(a), where the image resolution is fixed.

\paragraph{Runtime Breakdown.}
We decompose Graph2Scene into four stages: (1) \textbf{VLM Planning and Graph Construction}, which converts the input prompt into a structured scene graph using Qwen3-VL-8B-Instruct; (2) \textbf{Warm-up and Attention Analysis}, which performs an unguided generation pass to identify attention leakage and overlap; (3) \textbf{Mask Construction}, which builds selective attention masks based on graph connectivity and attention statistics; and (4) \textbf{Final Image Generation}, which performs graph-guided diffusion. We report the execution time of each stage as well as the total end-to-end latency, and compare Graph2Scene with Regional Prompt, the strongest baseline in our comparison. Unlike Graph2Scene, Regional Prompt does not require a separate warm-up pass or attention analysis stage. Therefore, its corresponding entry is marked as ``--'' (not applicable) rather than zero.

\paragraph{Discussion.}
With image resolution held constant, Graph2Scene introduces additional overhead, requiring 1.35$\times$--1.52$\times$ the runtime of Regional Prompt, mainly due to the warm-up and attention analysis stage. However, this overhead brings clear quality gains as prompt complexity increases. For 800--1000 tokens, Graph2Scene achieves 1.73$\times$, 2.52$\times$, and 3.53$\times$ the existence, attribute, and relationship accuracy of Regional Prompt, respectively. Runtime also increases with image resolution, mainly due to warm-up and final generation.

\begin{table}[!htbp]
\centering
\caption{Runtime breakdown and compositional quality of Graph2Scene
and Regional Prompt under different prompt lengths and image resolutions.
All runtime results are reported in seconds and averaged over five runs.
The VLM-based metrics evaluate object existence, attribute binding,
and relationship accuracy. Superscript ratios in (a) are computed
relative to Regional Prompt.}
\label{supp:tab:runtime_breakdown}
\small
\setlength{\tabcolsep}{1.85pt}
\renewcommand{\arraystretch}{1.25}

\begin{tabular*}{\linewidth}{@{\extracolsep{\fill}}llcccccccc@{}}
\toprule
\multicolumn{10}{c}{\textbf{(a) Prompt Scaling}
($1024\times1024$, seconds)} \\
\midrule
\shortstack{Prompt\\Tokens}
& Method
& \shortstack{VLM\\Planning}
& \shortstack{Warm-up\\\& Analysis}
& \shortstack{Mask\\Construction}
& \shortstack{Final\\Generation}
& \shortstack{Total\\Runtime}
& Acc$_{\text{exist}}$ $\uparrow$
& Acc$_{\text{attr}}$ $\uparrow$
& Acc$_{\text{rel}}$ $\uparrow$ \\
\midrule

\multirow{2}{*}{200--400}
& Graph2Scene
& 28.42 & 22.85 & 0.34 & 35.32
& 86.93\ratioBad{1.37}
& 1.00\ratioNeutral{1.00}
& 0.88\ratioBad{0.88}
& 0.93\ratioGood{1.09} \\
& Regional Prompt
& 28.69 & -- & 0.09 & 34.76 & 63.54
& 1.00 & 1.00 & 0.85 \\

\multirow{2}{*}{400--600}
& Graph2Scene
& 39.44 & 24.07 & 0.35 & 37.42
& 101.28\ratioBad{1.35}
& 0.94\ratioNeutral{1.00}
& 0.89\ratioGood{1.01}
& 0.88\ratioGood{1.24} \\
& Regional Prompt
& 39.08 & -- & 0.09 & 36.00 & 75.17
& 0.94 & 0.88 & 0.71 \\

\multirow{2}{*}{600--800}
& Graph2Scene
& 41.37 & 25.40 & 0.38 & 39.88
& 107.03\ratioBad{1.38}
& 0.80\ratioNeutral{1.00}
& 0.74\ratioGood{1.19}
& 0.66\ratioGood{1.89} \\
& Regional Prompt
& 40.15 & -- & 0.10 & 37.44 & 77.69
& 0.80 & 0.62 & 0.35 \\

\multirow{2}{*}{800--1000}
& Graph2Scene
& 56.46 & 27.18 & 0.40 & 42.45
& 126.49\ratioBad{1.52}
& \textbf{0.78}\ratioStrong{1.73}
& \textbf{0.63}\ratioStrong{2.52}
& \textbf{0.53}\ratioStrong{3.53} \\
& Regional Prompt
& 43.86 & -- & 0.10 & 39.21 & 83.17
& 0.45 & 0.25 & 0.15 \\

\midrule
\multicolumn{10}{c}{\textbf{(b) Resolution Scaling}
(400--600 Prompt Tokens, seconds)} \\
\midrule
Resolution
& Method
& \shortstack{VLM\\Planning}
& \shortstack{Warm-up\\\& Analysis}
& \shortstack{Mask\\Construction}
& \shortstack{Final\\Generation}
& \shortstack{Total\\Runtime}
& \multicolumn{3}{c}{VLM Metrics} \\
\midrule

512$\times$512
& \multirow{4}{*}{Graph2Scene}
& \multirow{4}{*}{39.44}
& 6.27 & 0.18 & 8.22 & 54.11
& \multicolumn{3}{c}{--} \\

768$\times$768
& & & 12.65 & 0.21 & 17.75 & 70.05
& \multicolumn{3}{c}{--} \\

1024$\times$1024
& & & 24.07 & 0.35 & 37.42 & 101.28
& \multicolumn{3}{c}{--} \\

1280$\times$1280
& & & 39.12 & 0.60 & 65.94 & 145.10
& \multicolumn{3}{c}{--} \\

\midrule

512$\times$512
& \multirow{4}{*}{Regional Prompt}
& \multirow{4}{*}{39.08}
& -- & 0.05 & 7.56 & 46.69
& \multicolumn{3}{c}{--} \\

768$\times$768
& & & -- & 0.06 & 14.33 & 53.47
& \multicolumn{3}{c}{--} \\

1024$\times$1024
& & & -- & 0.09 & 36.00 & 75.17
& \multicolumn{3}{c}{--} \\

1280$\times$1280
& & & -- & 0.17 & 54.91 & 94.16
& \multicolumn{3}{c}{--} \\

\bottomrule
\end{tabular*}
\end{table}

\FloatBarrier
\section{Additional Qualitative Results}
\label{supp:app:longbench}
\begin{figure}[!htbp]
\centering
\includegraphics[width=0.95\linewidth,keepaspectratio]{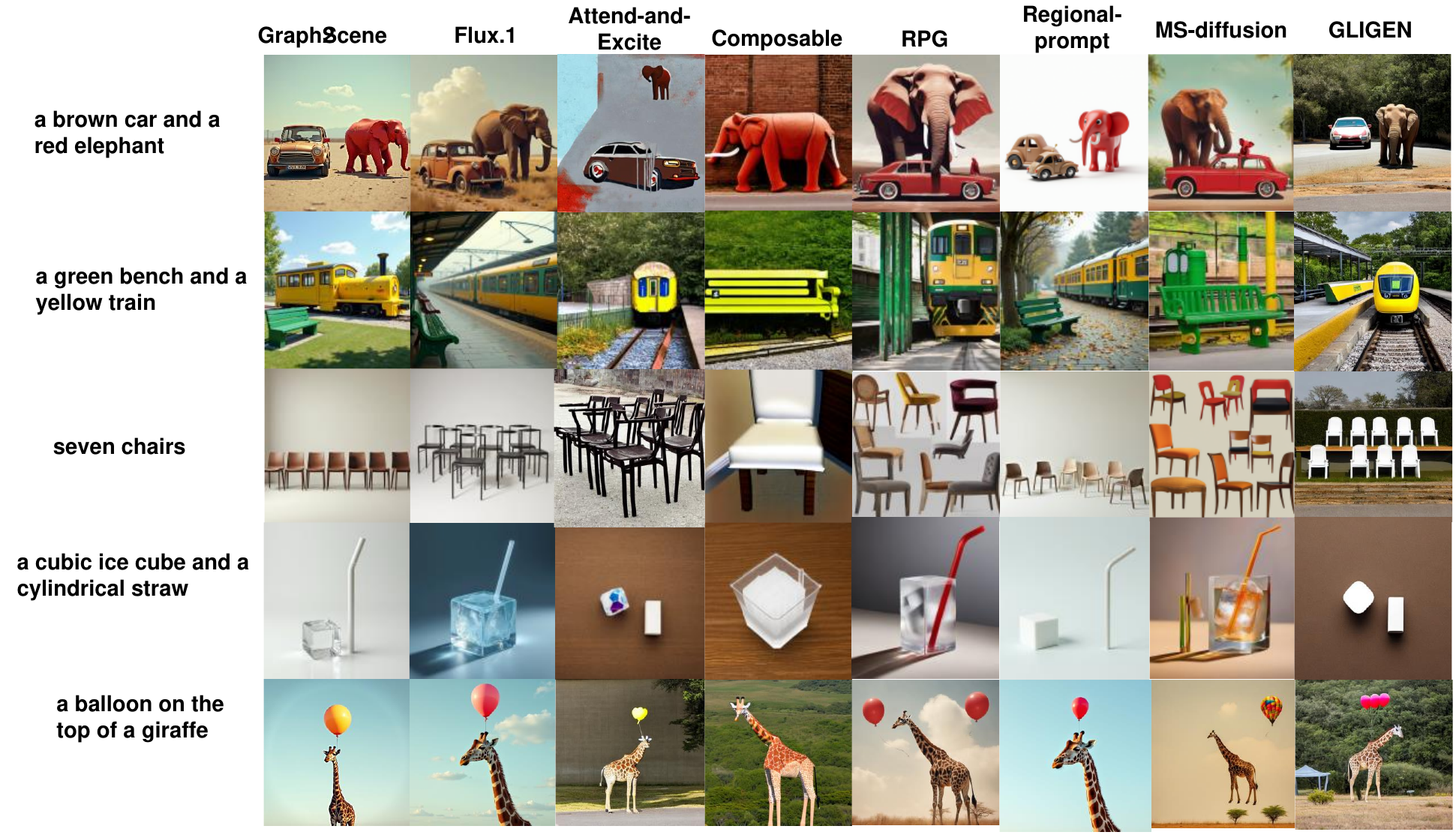}
\caption{Qualitative examples of Graph2Scene on simple scenes.}
\label{supp:fig:simple}
\end{figure}

\begin{figure}[p]
\centering
\includegraphics[width=0.95\linewidth]{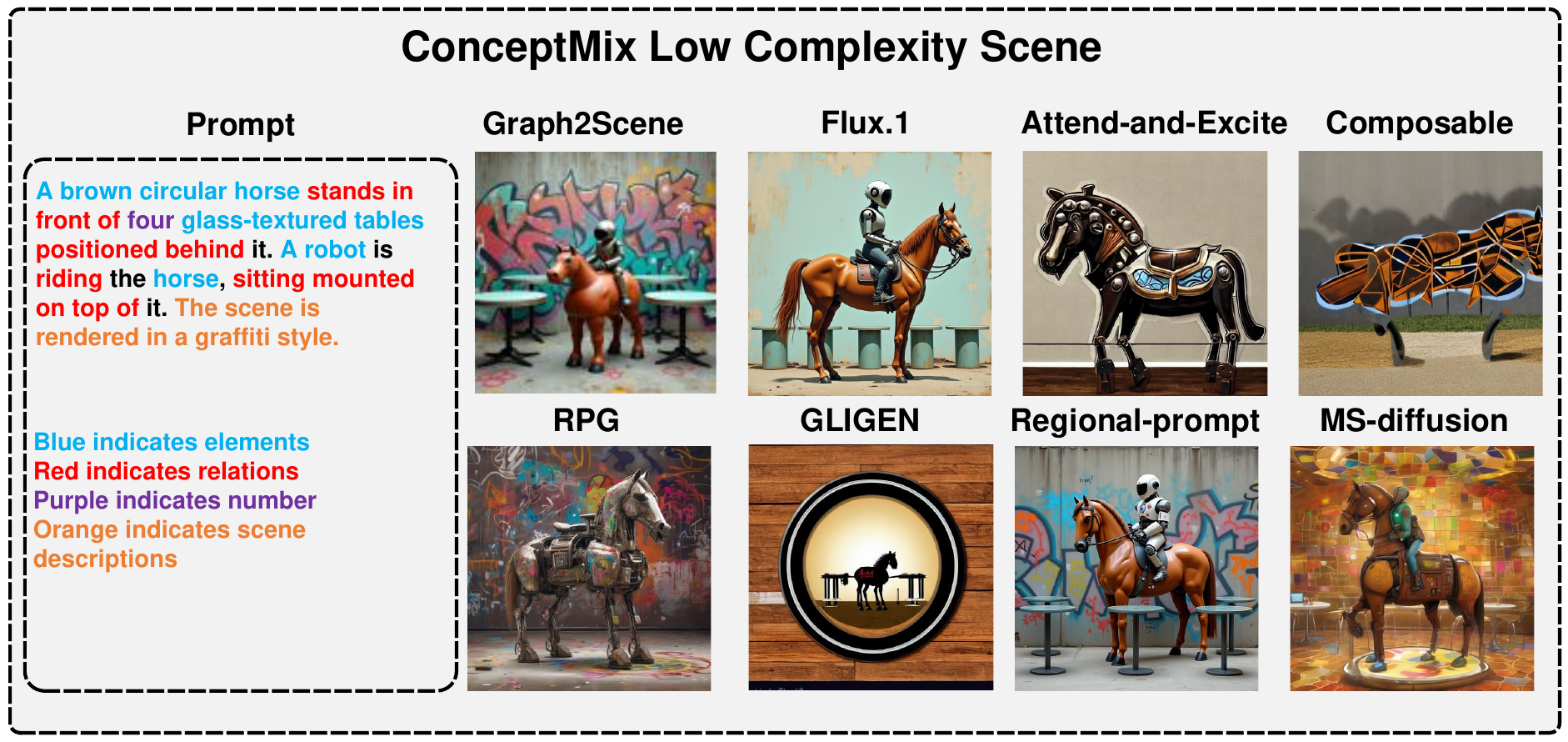}\par
\vspace{3pt}
\includegraphics[width=0.95\linewidth]{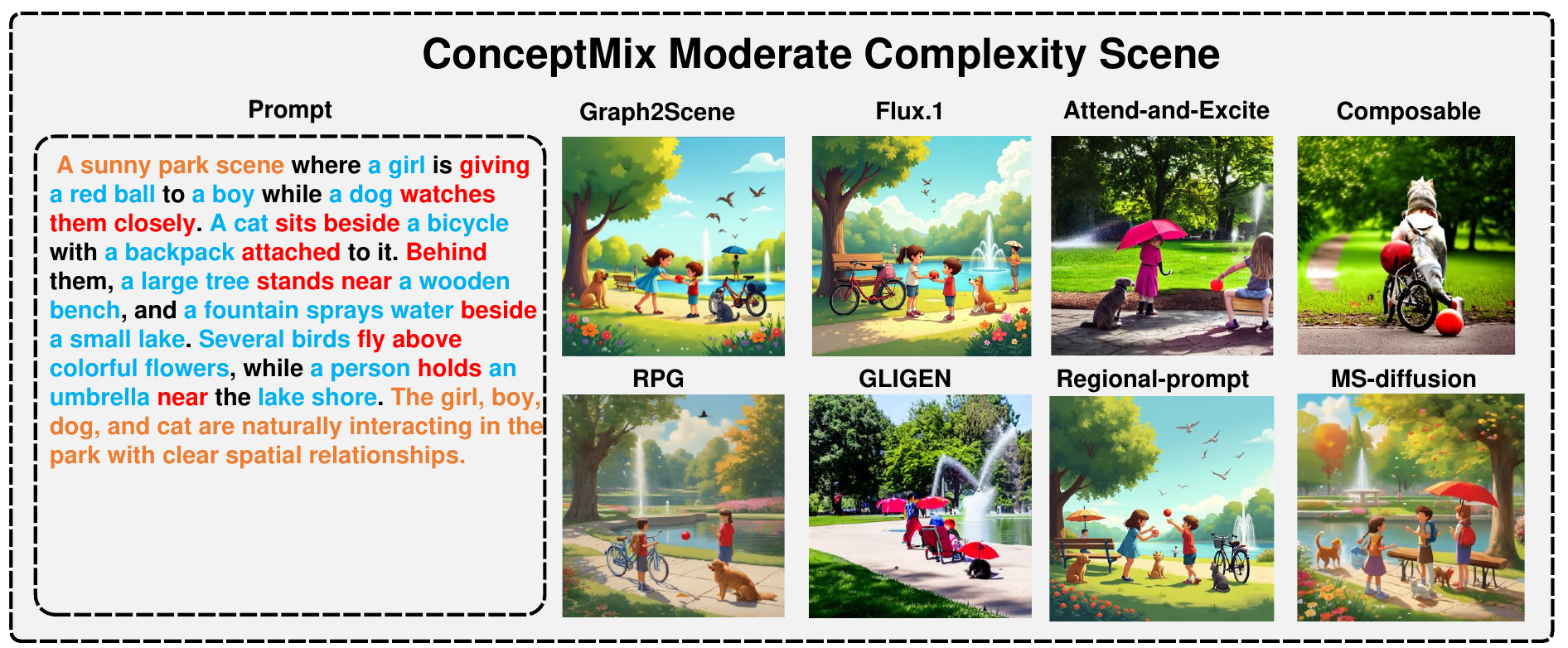}\par
\vspace{3pt}
\includegraphics[width=0.95\linewidth]{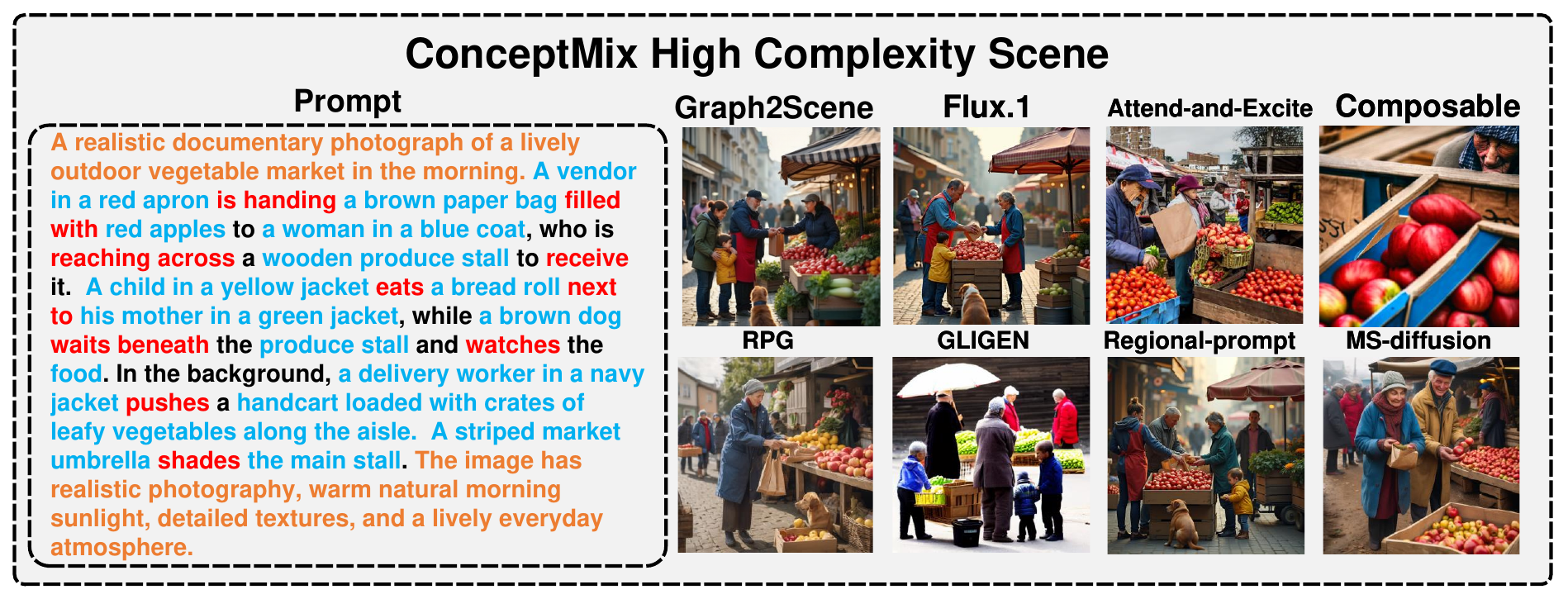}\par
\caption{Visualization examples on ConceptMix with increasing scene complexity.}
\label{supp:fig:conceptmix_complexity}
\end{figure}

\begin{figure}[p]
\centering
\includegraphics[width=0.85\linewidth]{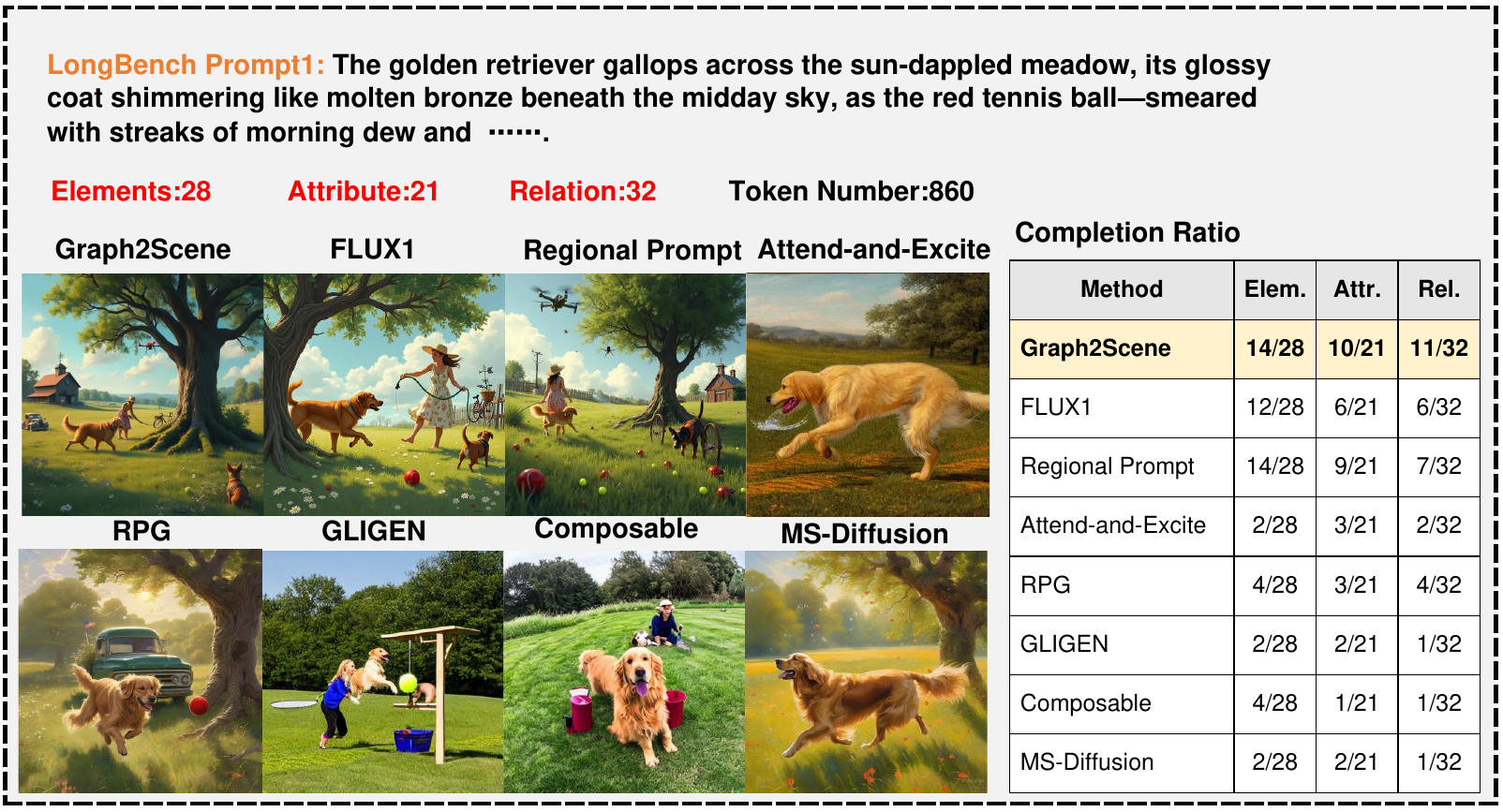}\par
\vspace{3pt}
\includegraphics[width=0.85\linewidth]{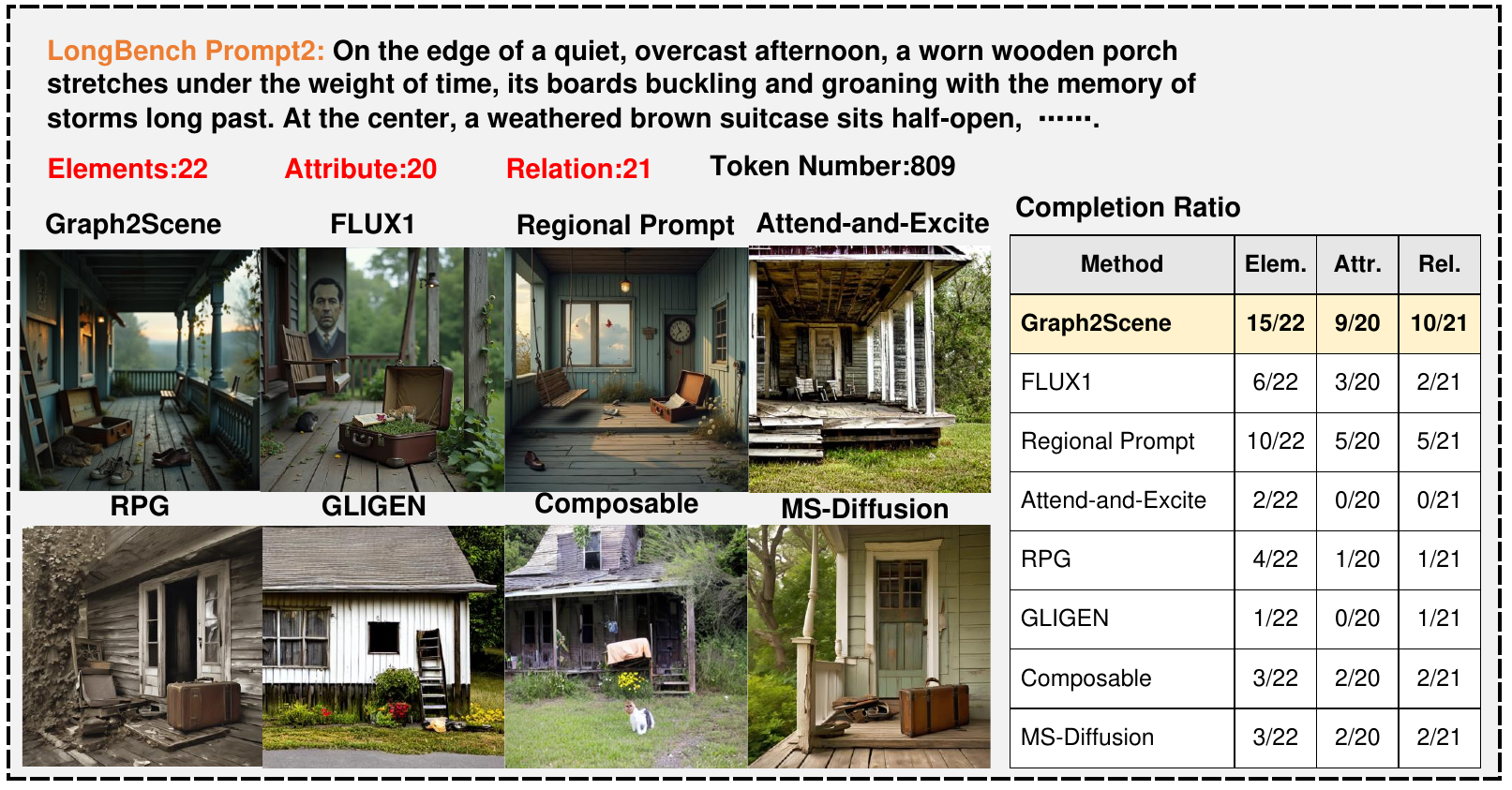}\par
\vspace{3pt}
\includegraphics[width=0.85\linewidth]{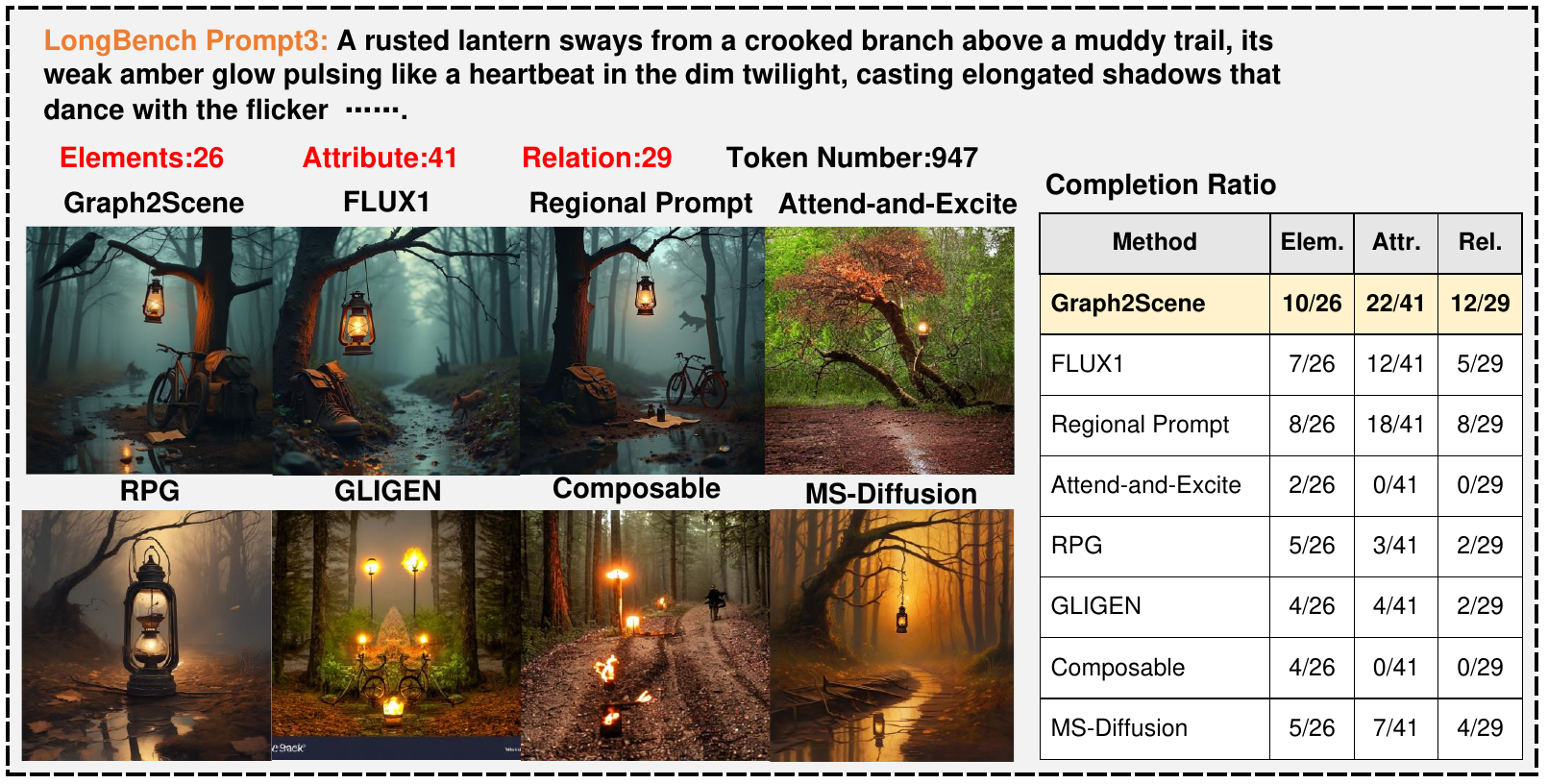}\par
\caption{Qualitative comparisons on LongBench-T2I with three different prompts.}
\label{supp:fig:longbench}
\end{figure}

We present qualitative results in three settings. Figure~\ref{supp:fig:simple} shows simple scenes with a small number of compositional constraints, while Figure~\ref{supp:fig:conceptmix_complexity} compares ConceptMix examples at low, moderate, and high scene complexity. Figures~\ref{supp:fig:longbench},~\ref{supp:fig:longbench_2}, and~\ref{supp:fig:longbench_3} extend the comparison to LongBench-T2I, whose long, highly compositional prompts test whether models can preserve multiple objects, attributes, and relationships in a single scene.

For LongBench-T2I, we use a VLM to evaluate the generated images in terms of object existence, attribute correctness, and relationship accuracy. Under these extremely long and complex prompts, all methods experience noticeable degradation: some required objects are omitted, attributes are incorrectly assigned or lost, and many relationships fail to appear in the generated scenes. This highlights the difficulty of faithfully following LongBench-T2I prompts with a large number of visual concepts and interactions. Nevertheless, Graph2Scene consistently preserves a larger proportion of the specified elements, attributes, and relationships than the compared methods. In particular, its advantage is more evident for attributes and relationships, indicating that explicitly organizing and controlling the interactions among scene elements helps maintain compositional consistency even when the prompt becomes highly complex.

\FloatBarrier
\clearpage
\clearpage

\FloatBarrier
\clearpage
\clearpage